\documentclass[journal]{IEEEtran}

\usepackage{graphics} 
\usepackage{epsfig} 
\usepackage{amsmath} 
\usepackage{amssymb}  
\usepackage{subfigure,cases,multirow}
\usepackage{algorithm,algpseudocode,algorithmicx}
\usepackage{amsfonts}
\usepackage[colorlinks,citecolor=blue,urlcolor=black,linkcolor=black]{hyperref}
\usepackage{accents}
\usepackage{mathrsfs}

\makeatletter
\def\wideubar{\underaccent{{\cc@style\underline{\mskip10mu}}}}
\def\ubar{\underaccent{{\cc@style\underline{\mskip6mu}}}}
\makeatother

\makeatletter
\def\widebar{\accentset{{\cc@style\underline{\mskip10mu}}}}
\makeatother

\ifCLASSOPTIONcompsoc
  \usepackage[nocompress]{cite}
\else
  \usepackage{cite}
\fi

\ifCLASSINFOpdf
\else
\fi
\def\rA{\mathscr A}
\def\cD{\mathcal D}
\def\cU{\mathcal U}
\def\cV{\mathcal V}
\def\cC{\mathcal C}

\def\cH{\mathcal H}
\def\cE{\mathcal E}
\def\cM{\mathcal M}
\def\cS{\mathcal S}
\def\cF{\mathcal F}
\def\cQ{\mathcal Q}
\def\cT{\mathcal T}
\def\cB{\mathcal B}

\def\km{\mathfrak m}
\def\kN{\mathfrak N}
\def\kq{\mathfrak q}
\def\kg{\mathfrak g}
\def\kp{\mathfrak p}
\def\kM{\mathfrak M}
\def\kC{\mathfrak C}
\def\kb{\mathfrak b}
\def\kL{\mathfrak L}

\def\bU{\mathbb U}
\def\bS{\mathbb S}
\def\bR{\mathbb R}
\def\bZ{\mathbb Z}

\def\hx{\hat{\mathbf x}}
\def\hq{\hat{\mathbf q}}
\def\hs{\hat{\mathbf s}}

\def\0{\mathbf 0}
\def\x{\mathbf x}
\def\q{\mathbf q}

\def\y{\mathbf y}
\def\s{\mathbf s}
\def\a{\mathbf a}
\def\b{\mathbf b}
\def\fu{\mathbf u}

\DeclareMathOperator\interp{I}
\DeclareMathOperator\Id{I_d}
\DeclareMathOperator\OF{\mathbf{OF}}

\def\<{\langle}
\def\>{\rangle}

\newlength\savewidth
\newcommand\shline{\noalign{\global\savewidth\arrayrulewidth\global\arrayrulewidth 1.0pt}\hline\noalign{\global\arrayrulewidth\savewidth}}

\newlength\savedwidth

\begin{document}
\title{Minimal Paths  for Tubular Structure Segmentation with Coherence Penalty and Adaptive Anisotropy}

\author{Da~Chen,~Jiong Zhang~and Laurent D. Cohen, ~\IEEEmembership{Fellow,~IEEE}
       
\thanks{Da~Chen and Laurent D. Cohen is with University Paris Dauphine, PSL Research University, CNRS, UMR 7534, CEREMADE, 75016 Paris, France. 
Da Chen is also with  Centre Hospitalier National d’Ophtalmologie des Quinze-Vingts, Paris, France. 
(e-mail: chenda@ceremade.dauphine.fr). (Da Chen and Jiong Zhang are the corresponding authors.)}
\thanks{Jiong Zhang is with the Laboratory of Neuro Imaging (LONI), USC Stevens Neuroimaging and Informatics Institute, Keck School of Medicine, University of Southern California, Los Angeles, CA 90033, USA. (e-mail: jiong.zhang@loni.usc.edu).}
}

\markboth{Journal of \LaTeX\ Class Files,~Vol.~14, No.~8, August~2018}
{Shell \MakeLowercase{\textit{et al.}}: Bare Demo of IEEEtran.cls for Journals}
\maketitle

\begin{abstract}
The  minimal path method has proven to be particularly useful and efficient in tubular structure  segmentation applications. In this paper, we propose a new minimal path model associated with a dynamic Riemannian metric embedded with an appearance feature coherence penalty  and an adaptive anisotropy enhancement term.  The features that characterize  the appearance and anisotropy properties of a tubular structure are extracted through the associated orientation score. The proposed dynamic Riemannian metric is updated in the course of  the geodesic distance computation carried out by the efficient single-pass fast marching method. Compared to state-of-the-art minimal path models,  the proposed minimal path model is able to extract the desired tubular structures from a complicated vessel tree structure. In addition, we propose an efficient prior path-based method to search for vessel radius value at each centerline position of the target. Finally, we perform the numerical experiments on both synthetic and real images. The quantitive validation is carried out on retinal vessel images. The  results indicate that the proposed model indeed achieves a promising performance.

\end{abstract}
\begin{IEEEkeywords}
Geodesic, appearance feature coherence, adaptive anisotropy,  dynamic metric, tubular structure segmentation.
\end{IEEEkeywords}
\IEEEpeerreviewmaketitle

\section{Introduction}
\label{sec:Intro}
Tubular structure segmentation plays an important role in many  applications of image analysis and medical imaging~\cite{kirbas2004review,lesage2009review,moccia2018blood}. A broad variety of significant approaches  have been exploited to solve the tubularity segmentation problem in the passed decades. Among these models, the variational methods including the active contours models (e.g.~\cite{lorigo2001curves,gooya2007effective,gooya2008r}) and the minimal path models~\cite{peyre2010geodesic} have been successfully applied to various situations thanks to their solid theoretical background and the reliable numerical solvers.

The basic idea for the active contours approaches is to model the boundaries of a tubular structure  through optimal curves or optimal surfaces, which  are in general  obtained by minimizing a  functional relying on the image features extracted from the tubular structures. An interesting example is the flux-based active contours model~\cite{vasilevskiy2002flux} using the spherical flux of the image gradient vectors for tubular features computation. Lax and Chung~\cite{law2009efficient} proposed an efficient way for the multi-scale spherical flux computation implemented  by the  fast Fourier transform in the Fourier domain. To improve the performance of the flux maximizing flow~\cite{vasilevskiy2002flux}, the image gradient vector field can be replaced by more adequate problem-dependent vector fields~\cite{law2007weighted,descoteaux2008geometric}.  
The authors of~\cite{law2010oriented} proposed a new active contours functional involving the asymmetry measure of the image gradients and the symmetric oriented flux measure~\cite{law2008three}. 
The constraints derived from the elongated nature of a tubular structure were taken into account in~\cite{nain2004vessel,gooya2008variational,manniesing2007vessel}.  Cohen and Deschamps~\cite{cohen2007segmentation} combined  the geodesic distance-based front propagation method~\cite{malladi1998real} and a Euclidean curve length-based thresholding scheme for 2D and 3D  vessel segmentation, by which the front leaking problem suffered by~\cite{malladi1998real} can be avoided in some extent.  Chen and Cohen~\cite{chen2018fast} generalized  that isotropic model~\cite{malladi1998real} to the anisotropic and asymmetric case through a Randers metric. 

It is important to indicate the other interesting and effective  models for tubular structure segmentation applications  including the curvilinear enhancement filters~(e.g. the steerable filters~\cite{sato1998three,frangi1998multiscale,law2008three,jacob2004design,moriconi2017vtrails}, the orientation score-based diffusion  filters~\cite{franken2009crossing,hannink2014crossing,zhang2016robust}, the path operator-based filter~\cite{merveille2018curvilinear}) and  the graph-based shortest path models~(e.g.~\cite{poon2007live,ulen2015shortest}). For more models relevant to  tubular structure segmentation, we refer to the complete reviews in~\cite{kirbas2004review,lesage2009review,moccia2018blood}. In the remaining of this section, we present  a non-exhaustive overview of the existing minimal path-based tubular structure segmentation approaches. 

The centerline of  a tubular structure  can be naturally  modelled as a minimal path~\cite{cohen1997global}, which is a globally  optimal curve that  minimizes a curve length measured by a suitable metric. The classical Cohen-Kimmel minimal path  model~\cite{cohen1997global} has been taken as the basic tool in many tubularity segmentation tasks,  due to its  global optimality  and   the efficient and stable  numerical solvers like the fast marching methods~\cite{sethian1999fast,mirebeau2014anisotropic,mirebeau2014efficient}. In the context of tubularity segmentation, the minimal path-based approaches are studied  mainly along two research lines. Firstly, the Cohen-Kimmel model~\cite{cohen1997global} provides an efficient and robust way for minimally interactive segmentation, providing that the end points of the target structure have been prescribed. 
To reduce the user intervention, the growing minimal path model~\cite{benmansour2009fast}  was designed to  iteratively add new source points, which are referred as keypoints, during the geodesic distance computation.  The keypoints detection method has been applied to road crack detection~\cite{kaul2012detecting} and blood vessel segmentation~\cite{li20093d,chen2016vesselkeypoints} with suitable stopping criteria. 
The geodesic voting model~\cite{rouchdy2013geodesic} used a voting score derived from a set of minimal paths with a common source point, which can detect a vessel tree structure from a single source point. The curves resulted  from the geodesic voting method~\cite{rouchdy2013geodesic} and their respective offset curves  can be taken as initialization for the narrowband active contours model~\cite{mille2009deformable}. By minimizing an adequate active contours energy, the results of~\cite{mille2009deformable} are able to depict the tubular centerlines and boundaries simultaneously.
In~\cite{chen2016curve}, the authors introduced a new curvilinearity extraction model  through a truncated  back-tracked geodesic scheme and the variant of the  geodesic voting score. The minimal path technique is often applied as the post-processing procedure in some vessel segmentation applications. In this case, the minimal paths can be used to get a connected vessel tree via a perceptual grouping way~\cite{cohen2001grouping,moriconi2017vtrails}.

Secondly, the metrics used in the original minimal path model~\cite{cohen1997global} cannot ensure that the geodesic paths will always pass through the exact tubular centerlines. Deschamps \emph{et al.}~\cite{deschamps2001fast} proposed a Euclidean distance-based potential construction method, where the geodesic paths can follow the tubular centerlines.  Li and Yezzi~\cite{li2007vessels}  defined an isotropic metric  over a multi-scale space to  simultaneously seek the  centerline and the boundary of the tubular structure. Benmansour and Cohen~\cite{benmansour2011tubular} generalized  this  isotropic model  to an anisotropic Riemannian case, where the vessel geometry  was extracted by the oriented flux filter~\cite{law2007weighted}. Moriconi~\emph{et al.}~\cite{moriconi2017vtrails} proposed a new tubular geometry descriptor based on a series of elongated Gaussian kernels for the multi-scale anisotropic Riemannian metric construction.
 P\'echaud~\emph{et al.}~\cite{pechaud2009extraction} added an abstract orientation dimension to the multi-scale space, which provides an orientation-lifted way to use the tubular anisotropy. 
 
The minimal path models mentioned above only consider the local vessel geometry to construct their metrics,  sometimes leading to a short combination problem or a shortcut problem. To solve these problems, the curvature regularization was taken  into account for geodesic computation  ~\cite{chen2017global,bekkers2015pde,mashtakov2017tracking,duits2018optimal,mirebeau2018fast}. This is based on the Eikonal equation framework, where the used metrics are commonly established in an orientation-lifted space. By the curvature-penalized fast marching method~\cite{mirebeau2018fast}, the geodesic paths associated to the Finsler elastica metric~\cite{chen2017global}, the sub-Riemannian metric~\cite{duits2018optimal}  can be efficiently estimated with adequate relaxations. Instead of using orientation lifting,  Liao~\emph{et al.}~\cite{liao2018progressive} proposed  a front frozen scheme for geodesic computation. They estimated a path feature from each point passed by the fast marching front, and froze these points  which do not satisfy the prescribed  criteria. However, these curvature-relevant models fail to exploit  the tubular appearance  coherence penalty which is an important property in many curvilinear structure  segmentation applications such as  retinal vessel segmentation and neural fibre extraction. 

In this paper, we propose a new metric penalized by the tubular appearance feature coherence measure, where the appearance features are characterized by coherence-enhanced orientation scores.  We estimate the feature coherence penalty during the geodesic distance computation in conjunction with a  truncated geodesic path tracking scheme~\cite{liao2018progressive,chen2016curve}. Thus the proposed metric is constructed in a dynamic manner.  In addition, we also propose a region-constrained metric established in a multi-scale space. The constrained region is the tubular neighbourhood of a prescribed curve,  yielding a radius-lifted geodesic path to depict  the target. Based on this metric, one can get a geodesic path involving both the centerline and the respective vessel thickness. The proposed method is very efficient and effective for single vessel extraction, especially when the target is  weakly-defined and close to a strong one. 
This document is an extension to the short  conference paper~\cite{chen2015interactive}, regarding which more contributions have been added.

\noindent\emph{Outline}.
This paper is organized  as follows: Section~\ref{sec:background} introduces the background on the tubular feature extractor and the tubular minimal path models. The construction of the appearance feature coherence penalized metric is introduced in Section~\ref{sec:MainWork}. The numerical implementation  is presented in Section~\ref{sec:FMImplementation}. The experimental results and the conclusion are presented in Sections~\ref{sec:Experiments} and \ref{Sec:Conclusion}, respectively.

\section{Finding Minimal Paths for Vessel Extraction}
\label{sec:background}   
\subsection{Tubular Feature Descriptor}
Let $\Omega\subset\bR^2$ be an open and   bounded domain instantiated in  2-dimension.
A multi-scale space is defined as a radius-lifted domain $\hat\Omega:=\Omega\times \bR_{\rm scale}\subset\bR^3$, where $\bR_{\rm scale}=[\Re_{\rm min}, \Re_{\rm max}]$ is a radius space. A point $\hx=(\x,r)\in\hat\Omega$  is a pair comprised of a position $\x\in\Omega$ and a radius value $r\in \bR_{\rm scale}$.

Without loss of generality, we assume that the gray levels inside the vessel are  locally darker than the background. 
A tubular structure  can be described  by the anisotropy feature vectors (vessel directions) and the   appearance features in the radius-lifted domain $\hat\Omega$.  
These features can be efficiently extracted by the steerable   filters such as~\cite{frangi1998multiscale,jacob2004design,law2008three}. We choose the optimally oriented flux~(OOF) filter~\cite{law2007weighted} as our vessel geometry detector. The $2$-dimentioanl OOF filter invokes a Gaussian kernel $G_\sigma$ with variance $\sigma$ and  a set of  circles with different radii. Let $\mathbf 1_{r}$ be an indicator function of a circle with radius $r$, which can be expressed by:
\begin{equation*}
\mathbf 1_{r}(\x)=
\begin{cases}
1,\quad &\text{if~}\|\x\|<r,\\
0,&\text{otherwise}.
\end{cases}	
\end{equation*}
The response $\OF$ of the OOF filter can be written by
\begin{equation}
\label{eq:OOF}
\OF(\x,r)=	\frac{1}{r} \big(\{\partial_{x_i x_j}G_\sigma\}_{\rm i,j}\ast \mathbf{1}_r\ast I\big)(\x),
\end{equation}
where $I:\Omega\to\bR$ is a scalar-valued image and the matrix  $\{\partial_{x_i x_j}G_\sigma\}_{\rm i,j}$ is the Hessian matrix of the Gaussian kernel $G_\sigma$ with $\partial_{x_i x_j}G_\sigma$  the second-order derivative along the axes $x_i$ and $x_j$.  For each point $\hx=(\x,r)$, the response $\OF(\hx)$ is a symmetric matrix of size  $2\times 2$ with  eigenvalues $\hat\varrho_1(\hx)$ and $ \hat\varrho_2(\hx)$. 
We assume that $\hat\varrho_1(\hx)\leq \hat\varrho_2(\hx),\,\forall\hx$. The anisotropy feature at $\hx$ can be set as  the eigenvector $\hat\kq_{\rm of}(\hx)$ of the matrix $\OF(\hx)$ corresponding to the eigenvalue $\hat\varrho_1(\hx)$. One can decompose the OOF response $\OF(\hx)$ by
\begin{equation}
\label{eq:OOFDecom}
\OF(\hx)=\hat\varrho_1(\hx)\hat\kq_{\rm of}(\hx)\otimes\hat\kq_{\rm of}(\hx)+\hat\varrho_2(\hx)\hat\kq^\perp_{\rm of}(\hx)\otimes\hat\kq^\perp_{\rm of}(\hx),
\end{equation}
where $\hat\kq_{\rm of}^\perp$ is the orthogonal vector of $\hat\kq_{\rm of}$ and the operator $\otimes$ is defined as $\fu_1\otimes\fu_2=\fu_1\fu_2^T,\,\forall \fu_1,\fu_2\in\bR^2$. The optimal scale map $\zeta_{\rm scale}:\Omega\to\bR_{\rm scale}$ can be expressed by
\begin{equation}
\label{eq:OptimalScale}	
\zeta_{\rm scale}(\x)=\mathop{\arg\,\!\max}_{r\in\bR_{\rm scale}}\,\hat\varrho_2(\hx),
\end{equation}
which defines the radius that  a tubular structure should have at $\x$. Note that for a point $\x$ that is inside a vessel, the eigenvalues satisfy that $\hat\varrho_1(\x,\zeta_{\rm scale}(\x))\approx 0$ and $\hat\varrho_2(\x,\zeta_{\rm scale}(\x))\gg0$ due to the lower gray levels inside the vessel regions. In this case, the vector $\hat\kq_{\rm of}(\x,\zeta_{\rm scale}(\x))$ points to the vessel direction at $\x$.

\subsection{Anisotropic Tubular Riemannian Minimal Path Model}
\label{subsec:TubularRMP}
The tubular minimal path models including the isotropic case~\cite{li2007vessels} and the anisotropic extension \cite{benmansour2011tubular}, aim to minimize the curve length of a radius-lifted path $\hat\gamma(u)=(\gamma(u), \tau(u))$ with $\tau:[0,1]\to\bR_{\rm scale}$ a parametric function. In this case,  the curve $\gamma(u)$ serves as  the vessel centerline position and $\tau(u)$ represents the radius of the vessel at the  position $\gamma(u)$. 
  
Let $S^+_d$ ($d=2,\, 3$) be the set of the symmetric  positive definite matrices of size $d\times d$ and let $\kL([0,1],\hat\Omega)$ be the set of all the Lipschitz paths $\hat\gamma:[0,1]\to\hat\Omega$. In the anisotropic case~\cite{benmansour2011tubular}, the length of a  curve $\hat\gamma\in\kL([0,1],\hat\Omega)$ associated to a tensor field $\cM_{\rm scale}:\hat\Omega\to S^+_3$  can be measured by
\begin{equation}
\label{eq:AnisotropicCurveLength}
\cE(\hat\gamma)=\int_0^1 \sqrt{\langle\hat\gamma^\prime(t),\cM_{\rm scale}(\hat\gamma(t))\hat\gamma^\prime(t)\rangle}~dt,
\end{equation}
where $\hat\gamma^\prime(t)=d\hat\gamma(t)/ dt$ is the first-order derivative of $\hat\gamma$.
According to~\cite{benmansour2011tubular}, the tensor $\cM_{\rm scale}(\hx)$ can be written as
\begin{equation}
\label{eq:RadiusLiftedTensor}
\cM_{\rm scale}(\hx)=
\begin{pmatrix}
\cM_{\rm aniso}(\hx)&\0\\
\0 &P_{\rm scale}(\hx)
\end{pmatrix},
\end{equation}
where $P_{\rm scale}:\hat\Omega\to\bR^+$ is a scalar-valued function  defined by
\begin{equation}
\label{eq:RadiusLiftedPotential}
 P_{\rm scale}(\hx)=\beta_{\rm scale}\exp\left(\frac{1}{2}\alpha_{\rm aniso}\,(\hat\varrho_1(\hx)+\hat\varrho_2(\hx))\right),
\end{equation}
where $\alpha_{\rm aniso}\in\bR$ and $\beta_{\rm scale}\in\bR^+$ are two constants.
The  tensor field $\cM_{\rm aniso}$  can be decomposed  by
\begin{align}
\label{eq:TensorDescirbeVessels}
\cM_{\rm aniso}(\hx)=&\exp(\alpha_{\rm aniso}\,\hat\varrho_2(\hx))\,\hat\kq_{\rm of}(\hx)\otimes\hat\kq_{\rm of}(\hx)\nonumber\\
&+\exp(\alpha_{\rm aniso}\,\hat\varrho_1(\hx))\hat\kq_{\rm of}^\perp(\hx)\otimes\hat\kq_{\rm of}^\perp(\hx).
\end{align}

Given a source point $\hs$ and a target point $\hx\in\hat\Omega$, the geodesic curve $\hat\cC_{\hs,\hx}\in\kL([0,1],\hat\Omega)$ linking from $\hs$ to $\hx$ is a global minimizer to the curve length $\cE$, i. e., 
\begin{equation}
\label{eq:Geodesic}
\hat\cC_{\hs,\hx}=\mathop{\rm{arg\,min}}\limits_{\hat\gamma\in\kL({[0,1],\hat\Omega)}}	\{\cE(\hat\gamma);~\hat\gamma(0)=\hs,~\hat\gamma(1)=\hx\}.
\end{equation}
For tubular structure extraction, a point $\hat\cC_{\hs,\hx}(t)$ in the geodesic path $\hat\cC_{\hs,\hx}$ involves three components, where the first two coordinates delineate a centerline position while the last one describes the radius the tubular structure has at that position.

The geodesic distance map $\cD_{\hs}:\hat\Omega\to\bR^+_0$ associated to  the source point $\hs$ is defined by 
\begin{equation}
\label{eq:MinimalAM}
\cD_{\hs}(\hx)=\inf_{\hat\gamma\in\kL([0,1],\hat\Omega)} \{\cE(\hat\gamma);~\hat\gamma(0)=\hs,~\hat\gamma(1)=\hx\}.	
\end{equation}
We define a norm $\|\fu\|_M=\sqrt{\<\fu,\,M\fu\>}$ for any matrix $M\in S^+_d$, where $\<\fu_1,\fu_2\>$ denotes the Euclidean scalar product of two vectors $\fu_1,\,\fu_2\in\bR^d$.
The geodesic distance map $\cD_{\hs}$ is the unique viscosity solution to the Eikonal PDE
\begin{equation}
\label{eq:AnisotropicEikonalPDE}
\|\nabla\cD_{\hs}(\hx)\|_{\cM_{\rm scale}^{-1}(\hx)}=1, \quad\forall \hx\in\hat\Omega\backslash \{\hs\}, 
\end{equation}
with boundary condition $\cD_{\hs}(\hs)=0$.

Let $\bar\cC_{\hx,\hs}\in\kL([0,L],\hat\Omega)$ be a geodesic curve parameterized by its arc-length with $\bar\cC_{\hx,\hs}(0)=\hx$ and $\bar\cC_{\hx,\hs}(L)=\hs$,  where $L$ is the Euclidean curve length of $\bar\cC_{\hx,\hs}$. The geodesic $\bar\cC_{\hx,\hs}$ can be computed by  solving the gradient descent ordinary differential equation (ODE) on the  map $\cD_{\hs}$ such that $\bar\cC_{\hx,\hs}(0)=\hx$ and 
\begin{equation}
\label{eq:GradientDescentODE}
\bar\cC_{\hx,\hs}^\prime(v)=-\frac{\cM_{\rm scale}^{-1}(\bar\cC_{\hx,\hs}(v))\nabla\cD_{\hs}(\bar\cC_{\hx,\hs}(v))}{\|\cM_{\rm scale}^{-1}(\bar\cC_{\hx,\hs}(v))\nabla\cD_{\hs}(\bar\cC_{\hx,\hs}(v))\|}.
\end{equation}
The  geodesic curve $\hat\cC_{\hs,\hx}$ with $\hat\cC_{\hs,\hx}(0)=\hs$ and $\hat\cC_{\hs,\hx}(1)=\hx$ can be recovered  by reversing and reparameterizing  $\bar\cC_{\hx,\hs}$.

\subsection{Short Branches Combination and Shortcut Problems}
\label{subsec:ShortBCP}
\begin{figure}[t]
\centering
\subfigure[]{\includegraphics[height=2.8cm]{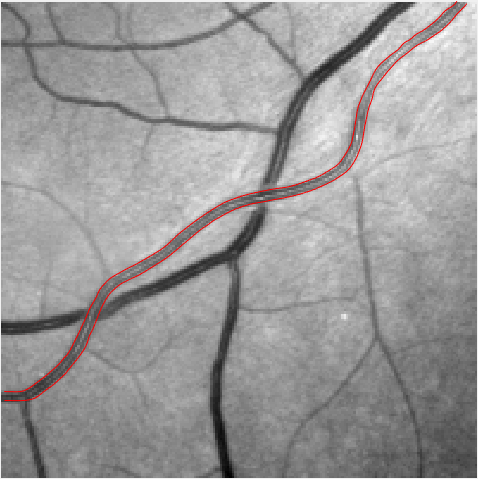}}~\subfigure[]{\includegraphics[height=2.8cm]{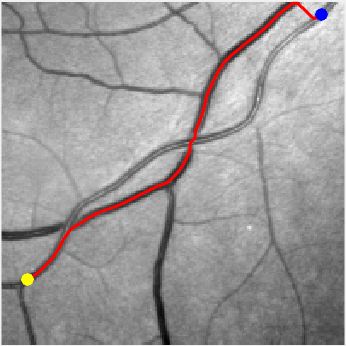}}~\subfigure[]{\includegraphics[height=2.8cm]{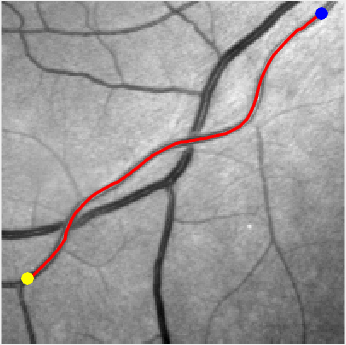}}
\caption{Short branches combination problem. (\textbf{a}) A retinal image  with red curves indicating the true boundaries of the target. (\textbf{b}) The minimal path  obtained from~\cite{benmansour2011tubular}. (\textbf{c}) The minimal path from the proposed model. The yellow and blue dots are the prescribed points.}
 \label{fig:ShortBranchesCombination}
 \end{figure}

The anisotropic tubular minimal path  model~\cite{benmansour2011tubular} invokes the static tensor field $\cM_{\rm scale}$ (see Eq.~\eqref{eq:RadiusLiftedTensor}), which relies on the pointwise  geometry features. When the target structure is weak in the sense of the appearance  feature and is close to or even crosses  a strong one, a minimal path derived from this model favours to pass through a way comprised of a set of strong vessel branches not belonging to the target. This gives rise to the short branches combination problem and the shortcut problem. 
In Fig.~\ref{fig:ShortBranchesCombination}, we make  use  of  a retinal image to illustrate these problems. In Fig.~\ref{fig:ShortBranchesCombination}a, the red curves  are the boundaries of the target vessel. This target appears to be weaker than its neighbouring vessel. In Fig.~\ref{fig:ShortBranchesCombination}b, we can see that the geodesic path derived from~\cite{benmansour2011tubular} passes by a long vessel segment not belonging to the target. In contrast, the proposed model which exploits  the appearance feature coherence penalization and the adaptive anisotropy enhancement is able to overcome  these problems as shown in Fig.~\ref{fig:ShortBranchesCombination}c.

\section{Dynamic Riemannian Metric with Appearance Feature  Coherence  Penalization}
\label{sec:MainWork}
\noindent\emph{Overview}.
 The main objective of this paper is to seek a tubular structure between two prescribed points from an image  involving a set of vessels, providing that the appearance\footnote{In general, the tubular appearance feature can be carried out by either the image gray levels, the vesselness or the orientation score.} and anisotropy features vary smoothly along the target structure. The tubular appearance features are supposed to be distinguishable between two structures close to each other. In this  setting, these tubular structures may  yield a set of crossing points, each of which  is defined as a point at the overlapped region.  We design a new metric by taking into account the  appearance feature coherence measure to overcome  the short branches combination problem. The existence of the crossing points may yield two discriminative appearance and anisotropy features in the overlapped region. In order to accurately compute the tubular appearance coherence measurement, we need to identify the correct appearance and anisotropy features belonging to the target structure at crossing points. For this purpose, we use the tool of the orientation score to extract the tubular appearance and  anisotropy features.
 
\subsection{Coherence-enhanced Orientation Score}
\label{subsec:SmoothedOS}
\noindent\emph{Asymmetrically Oriented Gaussian Kernels}.
Let $\bS^1=[0,2\pi)$ be an orientation space with periodic boundary condition and let  $\kg(\theta)=(\cos\theta,\sin\theta)^T$  be a unit  vector associated to an orientation $\theta\in\bS^1$. The  oriented Gaussian kernel~\cite{geusebroek2003fast} associated to an orientation $\theta\in\bS^1$ can be expressed by
\begin{equation*}
\cQ^\theta_{\kb}(\x)=\frac{1}{2\pi\sigma_1\sigma_2}\exp\left(\frac{|\<\kg(\theta),\x\>|^2}{-2\sigma_1^2}+\frac{|\<\kg^\perp(\theta),\x\>|^2}{-2\sigma_2^2}\right),
\end{equation*}
where $\kb=(\sigma_1,\sigma_2)$ includes the variances  $\sigma_1$, $\sigma_2\in\bR^+$ with $\sigma_1\gg\sigma_2$. The oriented Gaussian kernel $\cQ^\theta_{\kb}$ is symmetric with respect to the orientation $\theta$, i.e.,  $\cQ^\theta_{\kb}(\cdot)=\cQ^{\pi+\theta}_{\kb}(\cdot)$. 

\begin{figure}[t]
\centering		
\includegraphics[width=2.6cm]{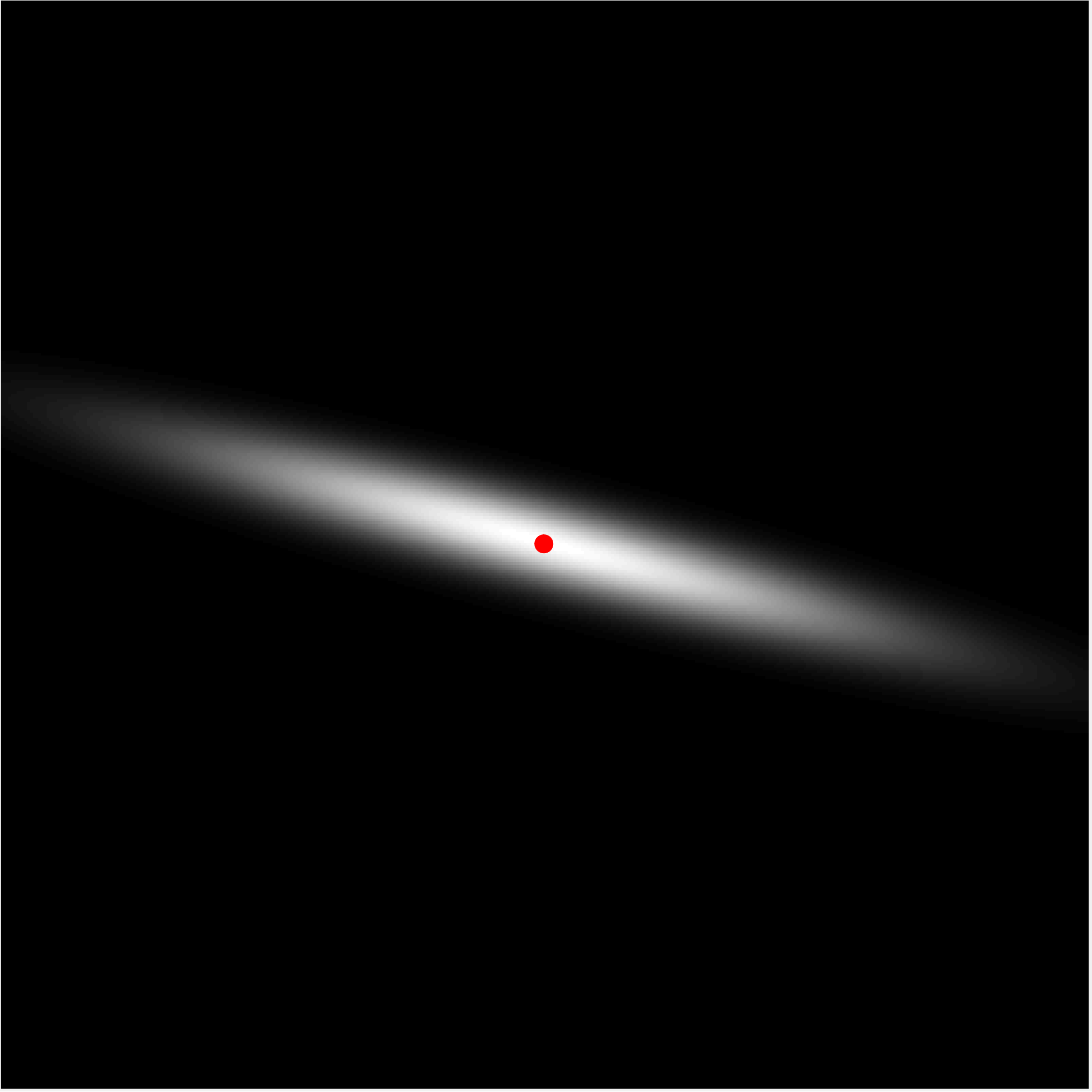}~\includegraphics[width=2.6cm]{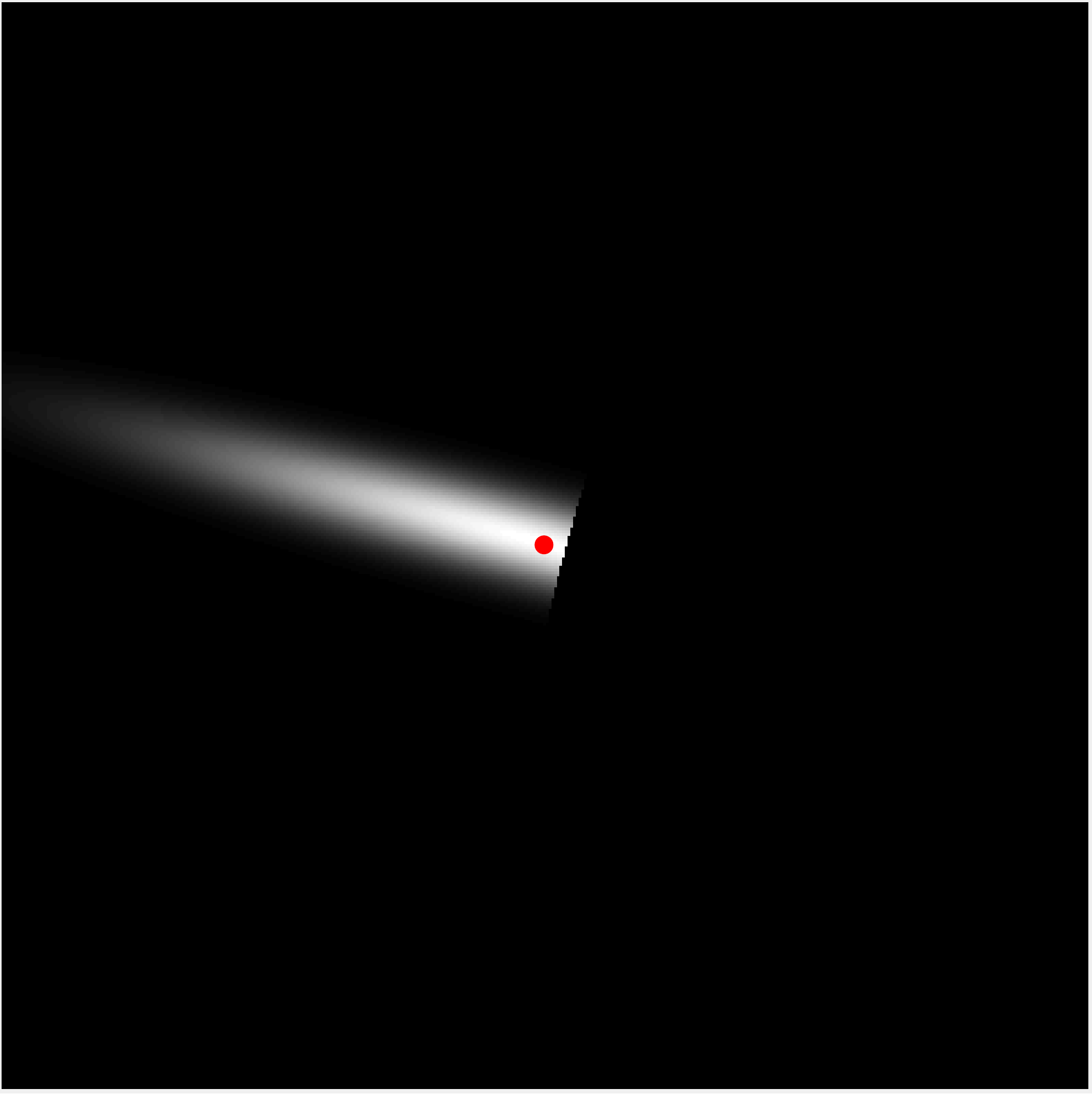}~\includegraphics[width=2.6cm]{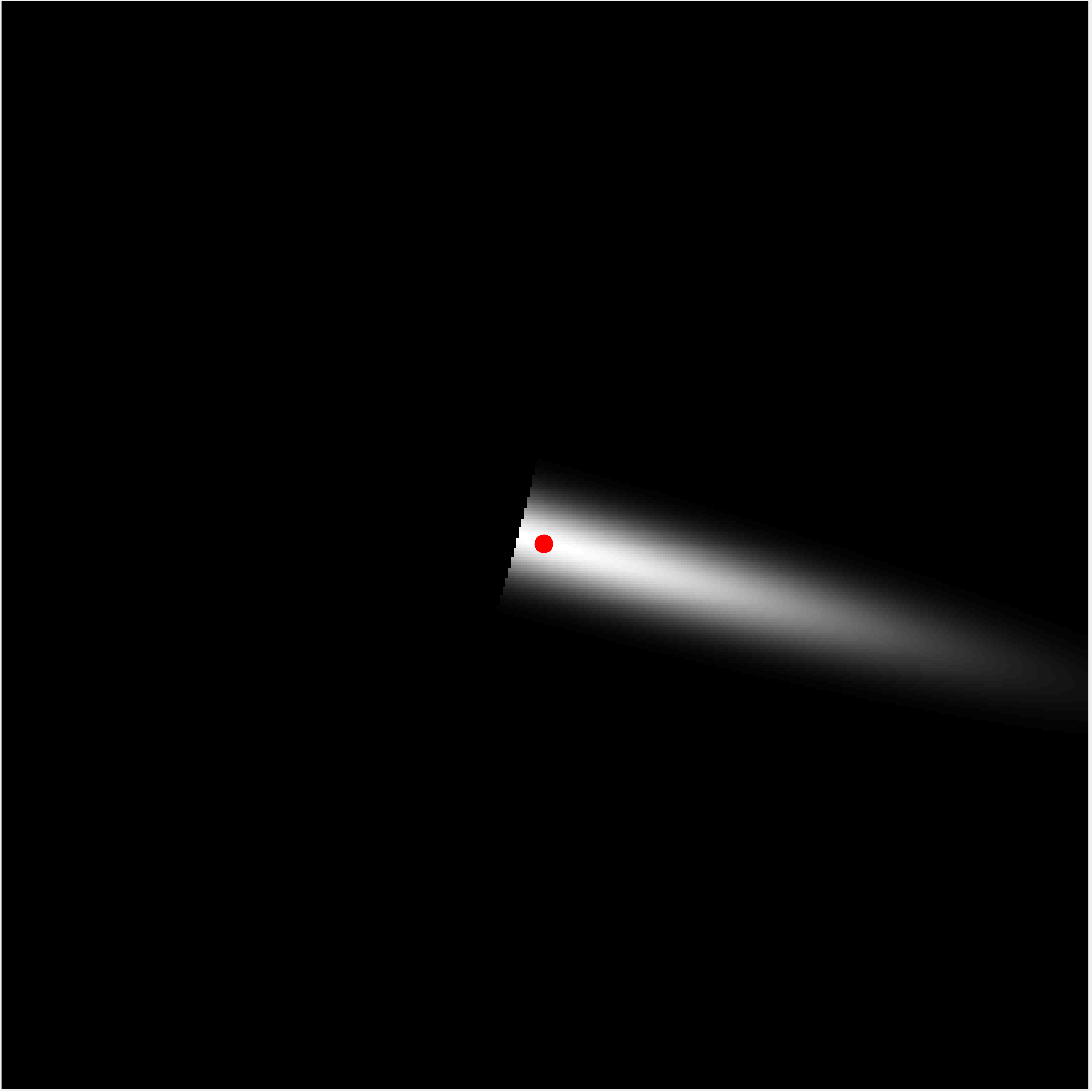}
\caption{A symmetric oriented Gaussian kernel (column 1) and the corresponding asymmetric kernels (columns 2-3). The red dots are the kernel centers.}
\label{fig:Kernels}
\end{figure}

We further consider an asymmetrically oriented Gaussian kernel relying on  a cutoff function $\delta^\theta:\Omega\to\{0,1\}$ such that 
\begin{equation}
\delta^\theta(\x)=
\begin{cases}
1,\quad&\text{if~}\<\nabla G_{\sigma_1}(\x),\,\kg(\theta)\> \geq \varepsilon_1,\\
0,&\text{otherwise},	
\end{cases}
\end{equation}
where  $\nabla G_{\sigma_1}=(\partial_{x_1} G_{\sigma_1}, \partial_{x_2} G_{\sigma_1})^T$ is the gradient of  a  Gaussian kernel $G_{\sigma_1}$ with variance $\sigma_1$ and $\varepsilon_1\approx0$ is a  constant. An asymmetric oriented Gaussian kernel $\cH^\theta_{\kb}$ associated to an orientation $\theta$ can be expressed by
\begin{equation}
\label{eq:AsyOrientedGuassian}
\cH_{\kb}^\theta(\x)=\delta^\theta(\x)\,\cQ^\theta_{\kb}(\x).
\end{equation}
In Fig.~\ref{fig:Kernels}, we illustrate an example  for the symmetric kernel $\cQ_\kb^\theta$  and the respective  asymmetric kernels $\cH_\kb^\theta$ and $\cH_\kb^{\theta+\pi}$. 

\begin{figure}[t]
\centering
\subfigure[]{\includegraphics[height=3.7cm]{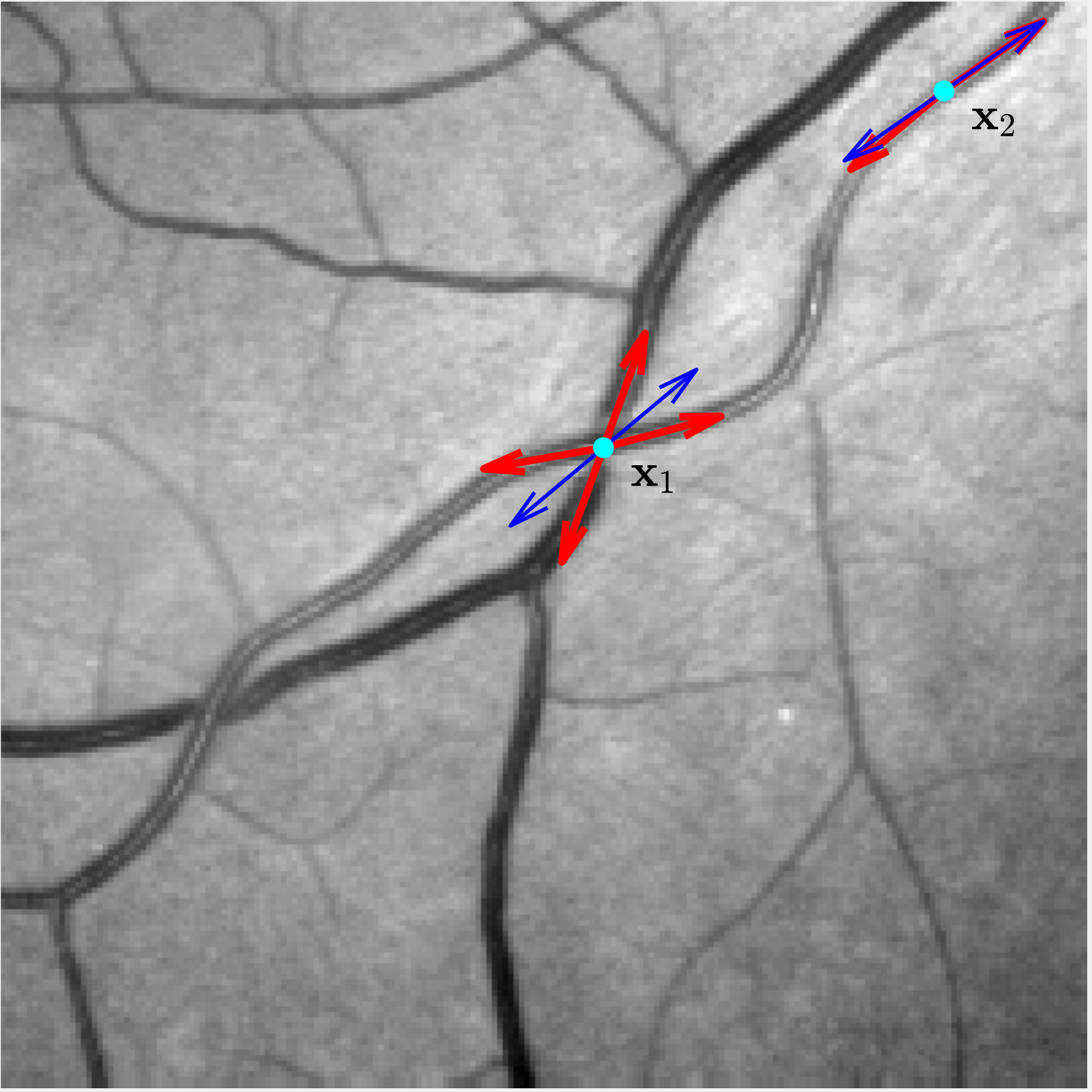}}~\subfigure[]{\includegraphics[height=3.7cm]{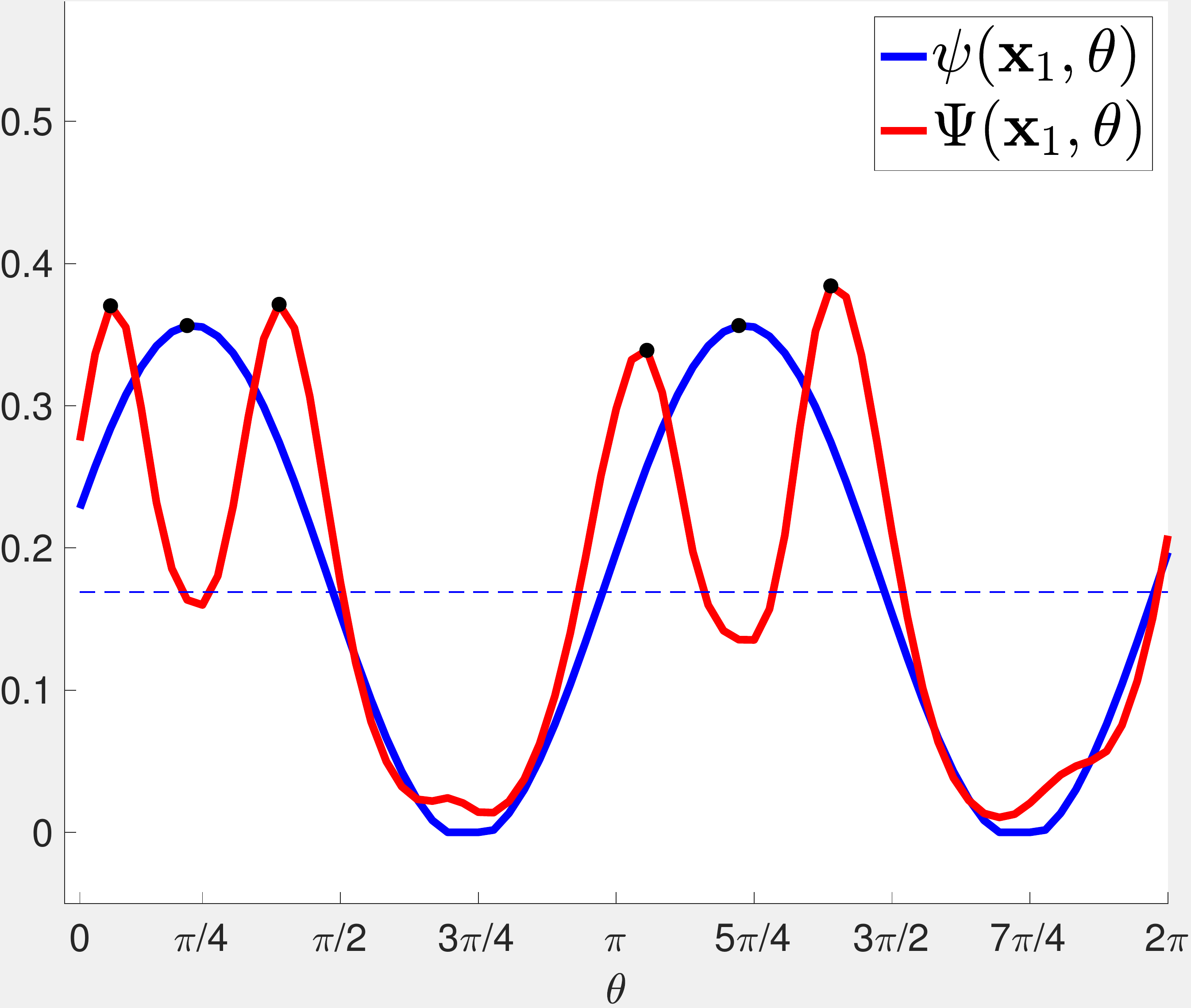}}
\caption{(\textbf{a}) A retinal image with two points $\x_1$ and $\x_2$. The red and blue arrows at  $\x_1$ (resp. $\x_2$) respectively indicate the elements involved in $\kM_{\x_1}$ (resp. $\kM_{\x_2}$) and $\kN_{\x_1}$ (resp. $\kN_{\x_2}$). (\textbf{b}) The red and blue curves respectively indicate the orientation scores of $\Psi(\x_1,\theta)$ and $\psi(\x_1,\theta)$ along the orientation dimention. The black dots indicate the peaks of $\Psi(\x_1,\cdot)$ and $\psi(\x_1,\cdot)$.}
\label{fig:SmoothOS}
\end{figure}

\noindent\emph{Coherence-enhancing orientation score}.
The orientation score  $\psi:\Omega\times\bS^1\to\bR_0^+$  can be computed by
\begin{equation}
\label{eq:OS}
\psi(\x,\theta)=\max\{\<\kg^\perp(\theta),\OF(\x,\zeta_{\rm scale}(\x))\kg^\perp(\theta)\>,0\},
\end{equation}
where $\OF$ is the OOF response defined in Eq.~\eqref{eq:OOF} and $\kg^\perp(\theta)=(-\sin\theta,\cos\theta)^T$. The scalar value $\zeta_{\rm scale}(\x)$ denotes the optimal scale at the point $\x$ (see Eq.~\eqref{eq:OptimalScale}). The orientation score $\psi$ sometimes still offers incorrect responses at crossing points due to the complex structures there, which can be seen from Fig.~\ref{fig:SmoothOS}. In Fig.~\ref{fig:SmoothOS}a,  the blue arrows at the crossing point $\x_1$ indicate the optimal feature vector $\kg(\theta^*)$ and $-\kg(\theta^*)$ where $\theta^*=\arg\max_\theta\{\psi(\x_1,\theta)\}$. Unfortunately, one can see that the blue arrows in Fig.~\ref{fig:SmoothOS}a are not proportional to the directions at the crossing point $\x_1$. This can be solved  by convolving the orientation score $\psi$ through the kernels $\cH^\theta_\kb$ to obtain a coherence-enhanced  orientation score $\Psi:\Omega\times\bS^1\to\bR^+_0$ 
\begin{equation}
\label{eq:EnhancedOS}
\Psi(\x,\theta)=\frac{(\cH^{\theta+\pi}_{\kb}\ast\psi^\theta)(\x)}{\int_{\Omega}\cH^{\theta+\pi}_{\kb}(\x)\,d\x},\quad \psi^{\theta}(\cdot)=\frac{\psi(\cdot,\theta)}{\|\psi\|_\infty},
\end{equation}
for each \emph{fixed} orientation $\theta\in\bS^1$, where $\ast$ is the convolution operator over the domain $\Omega$. The denominator $\int_{\Omega}\cH^{\theta+\pi}_{\kb}(\x)d\x$ in Eq.~\eqref{eq:EnhancedOS} is used for normalization.  

A set $\kM_\x$ of  locally optimal feature vectors with respect to the coherence-enhanced orientation score $\Psi$ can be defined by
\begin{equation}
\label{eq:LocalOrienMaximum}
\begin{split}
\kM_\x=\Big\{\kg(\theta^*);~&\Psi(\x,\theta^*)>\Psi(\x,\theta),~ \forall \theta\in N(\theta^*,\ell),\\
&\Psi(\x,\theta^*)>\frac{1}{2\pi}\int_0^{2\pi}\Psi(\x,\theta)\,d\theta\Big\},
\end{split}
\end{equation}
where  $N(\theta^*,\ell)$ denotes the interval of length $\ell$ centred at $\theta^*$. In Eq.~\eqref{eq:LocalOrienMaximum} we use the mean of $\Psi$ over the orientation dimension as  a thresholding value to identify the local maxima, which can be tuned adequately for different tasks.

An indicator $\kC_\x $  for the set $\kM_\x$ can be defined by
\begin{equation}
\label{eq:OptimalOrien}
\kC_\x(\theta)=
\begin{cases}
1,&\text{if~} \kg(\theta)\in\kM_\x,\\
0,&\text{otherwise}.
\end{cases}	
\end{equation}
Similar to $\kM_\x$, we can also define  a set $\kN_\x$ for each point $\x$ with respect to  the orientation score $\psi(\x,\cdot)$.

 In Fig.~\ref{fig:SmoothOS}, we show the advantages of the coherence-enhanced  orientation score $\Psi$ when comparing to the original $\psi$. In Fig.~\ref{fig:SmoothOS}a, the red  arrows at the crossing point $\x_1$ correspond to the feature vectors in the set $\kM_{\x_1}$ while the blue arrows at $\x_1$ indicate the feature vectors in the set $\kN_{\x_1}$.  We can see that the red arrows derived from $\Psi$ at $\x_1$ are approximately proportional  to the respective vessel directions,  while the blue arrows derived from $\psi$ point to incorrect directions. In Fig.~\ref{fig:SmoothOS}b, we plot the values of $\psi(\x_1,\theta)$ and $\Psi(\x_1,\theta)$ with respect to  $\theta$, where the black dots indicate the peaks of $\Psi(\x_1,\theta)$. In Fig.~\ref{fig:SmoothOS}a, the blue arrows at a non-crossing point $\x_2$ corresponding to the feature vectors in  $\kM_{\x_2}$ are almost propositional to the red arrows derived from $\kN_{\x_2}$. Each of the feature vectors indicated by the red and blue arrows well approximate  to  the respective vessel direction at $\x_2$.  
 
 \begin{figure}
 \centering
 \subfigure[]{\includegraphics[width=3cm]{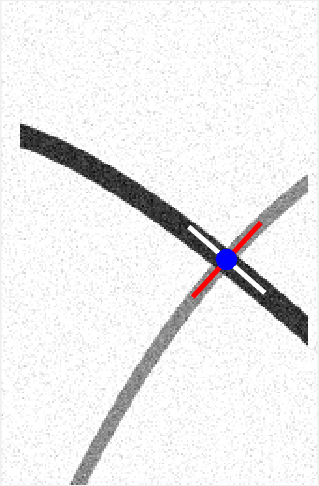}}~\subfigure[]{\includegraphics[width=3cm]{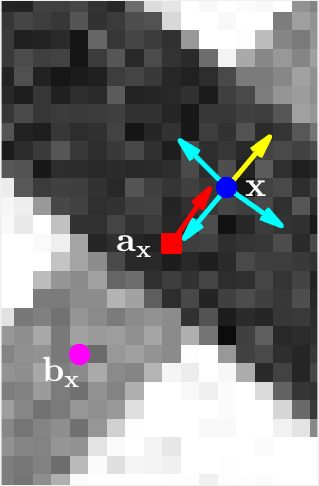}}
 \caption{\textbf{(a)} Blue dot indicates a crossing point $\x$ and the target tubularity has stronger gray levels. The red  line indicates the true  vessel direction. The white line is proportional to $\hat \kq_{\rm of}(\x,\zeta_{\rm scale}(\x))$. \textbf{(b)} Close-up view of the crossing region. The  arrows at $\x$ represent the vectors in $\kM_{\x}$ and the yellow one denotes $\kp(\x_{\rm n})$. The red arrow represents the vector $\kp(\a_\x)$ at the reference point $\a_\x$ (red square).  The magenta dot is the reference point $\b_\x$.}
 \label{fig:anisotropyVector}	
 \end{figure}

\subsection{A New Metric with Appearance Feature Coherence Penalty and Adaptive Anisotropy Enhancement}
\label{subsec:DynamicMetrics}
In this section, we  propose a new anisotropic metric $\cF_{\rm coh}:\Omega\times\bR^2\to[0,\infty]$ based on the appearance feature coherence penalty and the adaptive anisotropy enhancement. The metric $\cF_{\rm coh}$ is constructed based on a tensor field $\cT_{\rm coh}:\Omega\to S^+_2$
\begin{equation}
\label{eq:DynMetric}
\cF_{\rm coh}(\x,\fu)=\sqrt{\<\fu,\cT_{\rm coh}(\x)\fu\>}.	
\end{equation}
The tensor field $\cT_{\rm coh}$ is comprised of an appearance feature coherence penalty $\phi_{\rm coh}$ defined as a scalar-valued function, and two tensor fields $T_{\rm base}$ and $T_{\rm aniso}$. More precisely, it can be formulated  as 
\begin{equation}
\label{eq:DynamicTensor}
\cT_{\rm coh}(\x)=\phi_{\rm coh}(\x)\big(T_{\rm base}(\x)+T_{\rm aniso}(\x)\big),
\end{equation} 
The remaining of this section will be devoted to the computation of these components involved in the tensor field $\cT_{\rm coh}$.

We first present the computation methods respectively for the scalar-valued function $\phi_{\rm coh}$ and the tensor field $T_{\rm aniso}$. Both of them rely on a new feature vector field $\kp:\x\in\Omega\mapsto\kp(\x)\in\kM_{\x}$, which characterizes the anisotropy features  of the target tubular structure. In other words, the feature vector  $\kp(\x)$  is proportional  to the vessel direction at $\x$. 
For convenience, we define a function $\mu:\Omega\to\bS^1$ associated to the vector field $\kp$ being such that
\begin{equation}
\label{eq:CoupledOrienMap}
\big(\cos\mu(\x),\sin\mu(\x)\big)^T=\kp(\x).
\end{equation}

\noindent\emph{Adaptive anisotropy feature vector field}. At a crossing point $\x$, each vector in the set $\kM_\x$ corresponds to a vessel direction.  We need to choose the correct feature vector, i.e., the feature vector $\kp(\x)$, for the target vessel from the set $\kM_\x$. Recall that  the vector field $\kp$ is supposed to vary slowly  along the the same structure. This means that we can seek  $\kp(\x)$ from the set $\kM_{\x}$ using  a \emph{reference} point $\a_\x$ which is close to $\x$,  providing that $\kp(\a_\x)$ has been known. Note that the detection of the reference points is presented in Section~\ref{subsec:SinglePF}. In Fig.~\ref{fig:anisotropyVector}b, we denote the point $\x$ by a blue dot and its reference point $\a_\x$ by a red square. The cyan and yellow arrows at $\x$ (blue dot) represent the elements in the set $\kM_\x$ and the red arrow indicates the vector $\kp(\a_\x)$. 

Along the same tubular structure, the slow-varying property of the vector field $\kp$ in principle yields the maximal value of  $|\<\kp(\x), \kp(\a_\x)\>|$  among all the elements in $\kM_\x$, i.e., $|\<\kp(\x), \kp(\a_\x)\>|\geq |\<\fu, \kp(\a_\x)\>|,\,\forall\fu\in\kM_\x$.   We  define  a set $\kM^*_\x\subseteq\kM_\x$ involving all the \emph{maximal} feature vectors by
\begin{equation}
\label{eq:optimalOrienCandidates}
\kM^*_\x=\{\mathbf{w}\in\bR^2;~\mathbf{w}=\mathop{\arg\,\!\max}_{\fu\in\kM_\x}|\<\fu,\,\kp(\a_\x)\>|\}.	
\end{equation}
Based on the set  $\kM^*_\x$, the feature vector   $\kp(\x)$ can be identified by $\kp(\x)=(\cos\mu(\x),\sin\mu(\x))^T$, where the orientation $\mu(\x)\in\bS^1$ is computed by\footnote{Note that if the set $\{\theta\in\bS^1;\kg(\theta)\in\kM^*_\x\}$ includes more than one elements, we assign the smallest one to $\mu(\x)$.}
\begin{equation}
\label{eq:AdmissOriens}	
\mu(\x)=\mathop{\arg\,\!\min}_{\theta:\kg(\theta)\in\kM^*_\x}|\Psi(\x,\theta)-\Psi(\a_\x,\mu(\a_\x))|.
\end{equation}
In Fig.~\ref{fig:anisotropyVector}b, we show an example for the computation of $\kp(\x)$, where the yellow arrow at $\x$ represents the identified vector $\kp(\x)$ from the set $\kM_\x$. The vector field  $\kp$ is constructed in a progressive way. The initialization for $\kp$ is set as $\kp(\s)=(\cos\mu(\s),\sin\mu(\s))^T$, where $\mu(\s)=\arg\max_{\theta}\Psi(\s,\theta)$ and  $\s$ is the  source point.  The progressive procedure for the computation of $\kp$ is carried out  during the fast marching front propagation which is detailed in Section~\ref{subsec:SinglePF}.

\noindent\emph{Appearance feature coherence penalty}.
Once the anisotropy feature vector $\kp(\x)$ (or the corresponding orientation $\mu(\x)$) is detected, we can compute the appearance feature coherence penalty $\phi_{\rm coh}$ based on the coherence-enhanced  orientation score $\Psi$ (see Eq.~\eqref{eq:EnhancedOS})  and a new reference point $\b_\x$  by 
\begin{equation}
\label{eq:SimilarityOS}
\phi_{\rm coh}(\x)=\exp\big(\lambda\,|\Psi(\x,\mu(\x))-\Psi(\b_\x,\mu(\b_\x))|\big),
\end{equation}
where $\lambda$ is a positive constant. When the point $\x$ and its reference point $\b_\x$ are located at the same vessel, the value of $\phi_{\rm coh}(\x)$ should be low according to  the slow-varying prior for the appearance features.

\noindent\emph{Adaptively anisotropic tensor field}. The anisotropic tubular minimal path model~\cite{benmansour2011tubular} uses $\cM_{\rm scale}$ (see Eq.~\eqref{eq:RadiusLiftedTensor}) to compute the geodesic distances, which  grow slowly along the directions $\hat\kq_{\rm of}(\cdot)$ inside the vessel regions. However, at some crossing point $\x$, the directions $\hat\kq_{\rm of}(\x,\zeta_{\rm scale}(\x))$ from the OOF filter are not always proportional to the direction of the target vessel. This can be seen from Figs.~\ref{fig:SmoothOS}a and \ref{fig:anisotropyVector}a. In Fig.~\ref{fig:anisotropyVector}a, the direction of the target vessel  at $\x$ (denoted by the blue dot) is indicated by a red line. The white line indicates $\hat\kq_{\rm of}(\x,\zeta_{\rm scale}(\x))$ which is almost orthogonal to the red line.  In order to get a metric with correct anisotropy enhancement, we consider a  tensor field $T_{\rm aniso}$ formulated by
\begin{equation}
\label{eq:anisoAdaptive}
T_{\rm aniso}(\x)=\xi_{\rm aniso}\,\kp^\perp(\x)\otimes\kp^\perp(\x),
\end{equation}
where $\xi_{\rm aniso}$ is a positive constant.

Finally, the term $T_{\rm base}$ in Eq.~\eqref{eq:DynamicTensor} is an image data-driven tensor field. It can be formulated in an isotropic form:
\begin{equation}
\label{eq:StaticTensor1}
T_{\rm base}(\x)=\exp(-\alpha\,\max_{\theta\in\bS^1}\,\psi(\x,\theta))\,\Id,
\end{equation}
where $\Id$ is the $2\times 2$ identity matrix and $\alpha\in\bR^+$ is a constant. In order to take advantages of the anisotropy enhancement, we use the orientation score-based tensor field construction method~\cite{franceschiello2018neuro} to build $T_{\rm base}$. This is done by replacing the identity matrix in Eq.~\eqref{eq:StaticTensor1} by a new tensor field $T^{-1}_{\rm os}$  which is the inverse of a positive definite symmetric tensor field $T_{\rm os}$ 
\begin{equation}
\label{eq:OrienScoreTensor}
T_{\rm os}(\x)=\frac{\int_0^{2\pi}\kC_\x(\theta)\Psi(\x,\theta)\kg(\theta)\kg(\theta)^Td\theta}{\max\{\varepsilon_2, \int_0^{2\pi}\kC_\x(\theta)d\theta \} }+\xi_{\rm ident}\Id,
\end{equation}
where  $\xi_{\rm ident}$ and $\varepsilon_2$ ($\varepsilon_2\approx0$) are two small positive constants. The matrix $\xi_{\rm ident}\Id$ ensures the non-singularity of the matrix  $T_{\rm os}(\x),\,\forall\x\in\Omega$. Increasing the value of $\xi_{\rm ident}$ may reduce the anisotropy property of $T_{\rm os}(\x)$. The desired tensor field $T_{\rm base}$ can be  computed by
\begin{equation}
\label{eq:StaticTensor2}
T_{\rm base}(\x)=\exp(-\alpha\,\max_{\theta\in\bS^1}\,\psi(\x,\theta))\,T^{-1}_{\rm os}(\x),
\end{equation}
where the parameter $\alpha\in\bR^+$  controls the influence from the image data. At the crossing points, the tubular appearance and anisotropy features derived from both of the two overlapped structures will contribute to the tensor fields $T_{\rm os}$ and $T_{\rm base}$. In the following experiments, we make use of the tensor field $T_{\rm os}$ defined in Eq.~\eqref{eq:OrienScoreTensor} to build the tensor field $T_{\rm base}$.

\begin{figure}[t]
\centering	
\subfigure[]{\includegraphics[height=3.8cm]{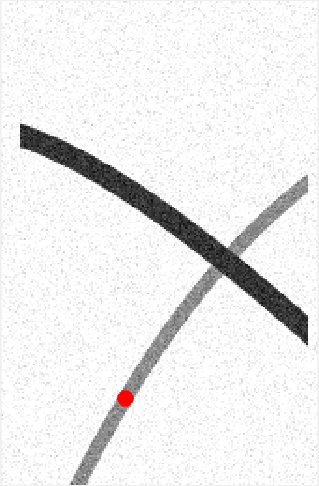}}~\subfigure[]{\includegraphics[height=3.8cm]{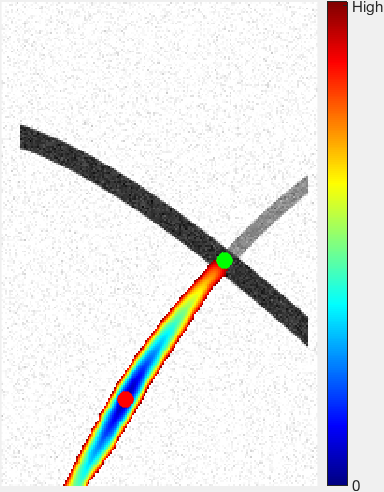}}~\subfigure[]{\includegraphics[height=3.8cm]{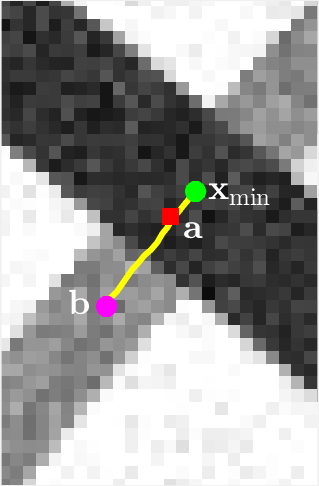}}
\caption{\textbf{(a)} Two tubular structures crossing each other. The red dot indicates the source point. \textbf{(b)} Geodesic distances superimposed on the synthetic image. The green dot is the latest \emph{Accepted} point $\x_{\rm min}$. \textbf{(c)} Close-up view of the region around $\x_{\rm min}$. The red square and the magenta dot are the reference points $\a$ and $\b$ respectively. The yellow line denotes the back-tracked short geodesic path.}
\label{fig:ReferencePoints}
\end{figure}

\section{Fast  Marching Implementations}
\label{sec:FMImplementation}

\subsection{Fast Marching Fronts Propagation Scheme}
\label{subsec:FMScheme}
In this section, we introduce the general scheme for the fast marching method which is first introduced in~\cite{sethian1999fast,tsitsiklis1995efficient}. It is an efficient way for the computation of the geodesic distances on the discretization domain $\bZ^2$ of the image domain $\Omega$. Basically, the fast marching fronts visit all the grid points in a monotonically increasing order expanding from a set of source points,  coupled with a course of label assignment operation through a map $\cV:\bZ^2\to\{$\emph{Far}, \emph{Accepted}, \emph{Front}$\}$. One of the crucial point of the fast marching method is the neighbourhood system used.  In contrast to the  isotropic fast marching method~\cite{sethian1999fast} which invokes an 4-connectivity neighbourhood system,  the anisotropic variant of the fast marching method\footnote{C++ codes: \href{https://github.com/Mirebeau/ITK_Anisotropic}{https://github.com/Mirebeau/ITK\_Anisotropic}.}~\cite{mirebeau2014anisotropic} used in this paper requires a more complicated metric-dependent neighbourhood $\cS$. It  utilizes the geometry tool of  the Lattice basis reduction  and  achieves an excellent balance between the  complexity and accuracy for the geodesic distance computation. For the sake of simplicity, we define an inverse neighbourhood $\cS^{-1}$ for each grid point $\x\in\bZ^2$ such that $\cS^{-1}(\x):=\{\mathbf z\in\bZ^2;\x\in \cS(\mathbf z)\}$. Thus a grid point $\y\in\bZ^2$ is a neighbour point of $\x$ if $\y\in\cS^{-1}(\x)$. We refer to~\cite{mirebeau2014anisotropic, mirebeau2014efficient} for more details on $\cS$ and $\cS^{-1}$.

Let $\cU_\s:\bZ^2\backslash\{\s\}\to\bR^+_0$ be the geodesic distance map associated to the metric $\cF_{\rm coh}$ defined in~\eqref{eq:DynMetric}, where $\s\in\bZ^2$ is the source point such that $\cU_\s(\s)=0$. In each geodesic  distance update iteration, a point $\x_{\rm min}\in\bZ^2$ with minimal $\cU_\s$ among all the \emph{Front} points can be detected by
\begin{equation}
\label{eq:LatestAccepted}
\x_{\rm min}=\mathop{\arg\,\!\min}_{\x:\cV(\x)=\emph{Front}}\cU_\s(\x).
\end{equation} 
The  point $\x_{\rm min}$ is immediately  tagged as \emph{Accepted}. In the following, $\x_{\rm min}$ is called the latest \emph{Accepted} point.
The geodesic distances for all the neighbour points $\x_{\rm n}\in\cS^{-1}(\x_{\rm min})$ with $\cV(\x_{\rm n})\neq$\emph{Accepted} can be  estimated by the solution to  the Hopf-Lax operator~\cite{mirebeau2014anisotropic}
\begin{equation}
\label{eq:HopfLax}
\cU_\s(\x_{\rm n})=\min_{\mathbf{z}\in\partial \cS(\x_{\rm n})}\{\cF_{\rm coh}(\x_{\rm n},\mathbf{z}-\x_{\rm n})+ \interp_{\cS(\x_{\rm n})}\cU_{\s}(\mathbf{z})\}
\end{equation}
where $\interp_{\cS(\cdot)}$ is a piecewise linear interpolator and $\interp_{\cS(\x_{\rm n})}\cU_{\s}(\mathbf{z})$ is a distance value estimated by the  interpolator $\interp_{\cS(\x_{\rm n})}$ in the neighbourhood $\cS(\x_{\rm n})$. The Hopf-Lax operator in Eq.~\eqref{eq:HopfLax} is an approximation to the Eikonal equation based on Bellman's optimality principle. 

\subsection{Single Front Propagation Implementation}
\label{subsec:SinglePF}
In this section, we present the method for updating the metric $\cF_{\rm coh}$ in conjunction with the detected reference points. This is done by the fast marching front propagation scheme~\cite{mirebeau2014anisotropic} and a truncated geodesic curve back-tracking scheme~\cite{liao2018progressive}. Since the metric $\cF_{\rm coh}$ is constructed during the front propagation, we refer to $\cF_{\rm coh}$ as a dynamic Riemannian metric. 

In each geodesic distance update iteration, we first search for the latest \emph{Accepted} point $\x_{\rm min}$ from all the \emph{Front} points. In Fig.~\ref{fig:ReferencePoints}b, we take a green dot as an example for such a point $\x_{\rm min}$. From $\x_{\rm min}$ we can track a geodesic path $\bar\cC_{\x_{\rm min}}$ by solving  the following  gradient descent ODE on $\cU_\s$  
\begin{equation}
\label{eq:physicalODE}
\bar\cC^\prime_{\x_{\rm min}}(v)=-\frac{\cT^{-1}_{\rm coh}(\bar\cC_{\x_{\rm min}}(v))\nabla\cU_\s(\bar\cC_{\x_{\rm min}}(v))}{\|\cT^{-1}_{\rm coh}(\bar\cC_{\x_{\rm min}}(v))\nabla\cU_\s(\bar\cC_{\x_{\rm min}}(v))\|},
\end{equation}
with $\bar\cC_{\x_{\rm min}}(0)=\x_{\rm min}$ and $\bar\cC_{\x_{\rm min}}(L)=\s$, where $L$ is the Euclidean curve length of $\bar{\cC}_{\x_{\rm min}}$.  Since each neighbour point $\x_{\rm n}\in\cS^{-1}(\x_{\rm min})$ is close to $\x_{\rm min}$, we can seek the reference points $\a_{\x_{\rm n}}$ and $\b_{\x_{\rm n}}$ for the points $\x_{\rm n}$ with $\cV(\x_{\rm n})\neq$\emph{Accepted} through the geodesic path $\bar\cC_{\x_{\rm  min}}$ as follows:
\begin{equation}
\label{eq:ReferencePoints}	
\a_{\x_{\rm n}}=\bar{\cC}_{\x_{\rm min}}(u_1),\quad \b_{\x_{\rm n}}=\bar{\cC}_{\x_{\rm min}}(u_2),	
\end{equation}
where $u_1,\,u_2\in(0,L)$ are two positive constants  and $u_1\leq u_2$. In this case, in each distance  update iteration, all the non-accepted neighbour points $\x_{\rm n}$ of $\x_{\rm min}$ share the same reference points. Henceforth we respectively denote by $\a$ and $\b$ the reference points $\a_{\x_{\rm n}}$ and $\b_{\x_{\rm n}}$  for simplicity. In Fig.~\ref{fig:ReferencePoints}c,  the portion of the geodesic path $\bar\cC_{\rm \x_{\rm min}}$ between $\x_{\rm min}$ (green dot) and the reference point $\a$ (red square) is illustrated  by a yellow line.  The reference points $\a$ and $\b$ for all the non-accepted neighbour points $\x_{\rm n}$ are respectively denoted by a red square and a magenta dot.

The fast marching front propagation scheme  provides a  progressive way to identify the reference points, which can be used to update the metric $\cF_{\rm coh}$. The details are illustrated in  Algorithm~\ref{algo:SingleFP}. In practice, the reference points  are detected via two thresholding values $\chi_1,\,\chi_2$ with $\chi_1\leq \chi_2$, which denote the numbers of grid points passed through by each back-tracked geodesic. This tracking processing is terminated once the reference point corresponding to the larger thresholding value $\chi_2$ is obtained or the source point is reached.

\begin{figure}[t]
\centering	
\subfigure[]{\includegraphics[width=3.6cm]{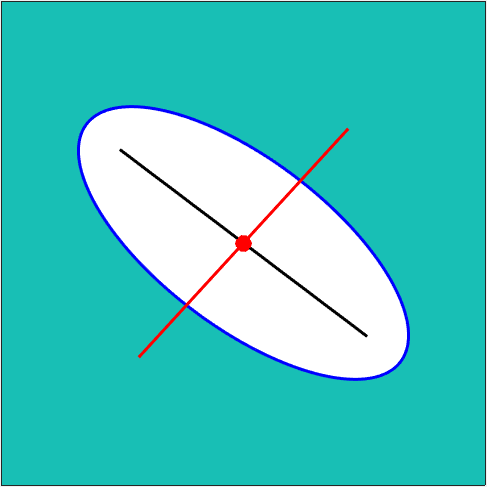}}~\subfigure[]{\includegraphics[width=3.6cm]{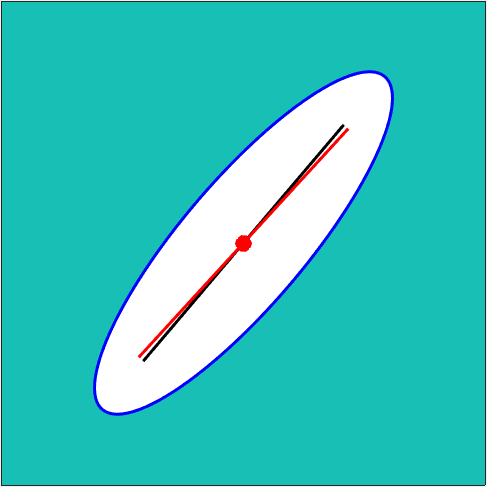}}
\caption{\textbf{(a)}-\textbf{(b)}: Control sets  $\cB_{\rm aniso}(\x_{\rm min})$ and $\cB_{\rm coh}(\x_{\rm min})$ derived from the picture in Fig.~\ref{fig:ReferencePoints}a, where $\x_{\rm min}$ is indicated by the green dot in Fig.~\ref{fig:ReferencePoints}b. In each figure, the black line indicates the direction of the major axis of the corresponding ellipse while the red line indicates the direction of the target vessel shown in Fig.~\ref{fig:ReferencePoints}a at the crossing point.}
\label{fig:ControlSet}
\end{figure}

\begin{figure*}[t]
\centering
\subfigure[]{\includegraphics[height=3.8cm]{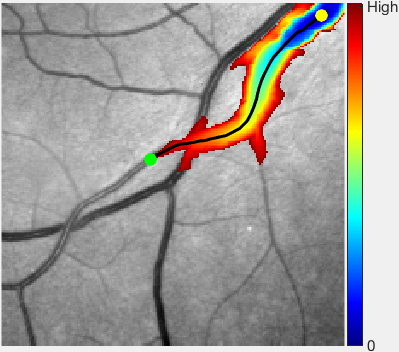}}~\subfigure[]{\includegraphics[height=3.8cm]{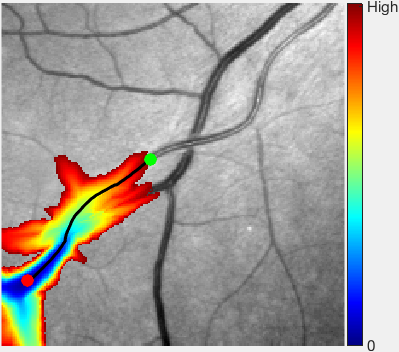}}
\subfigure[]{\includegraphics[height=3.8cm]{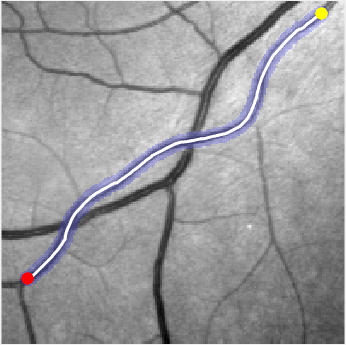}}~\subfigure[]{\includegraphics[height=3.8cm]{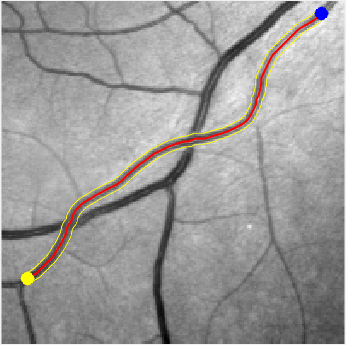}}
\caption{(\textbf{a}) The geodesic distance map $\cU_{\s}$ superimposed on the retinal patch. The black line indicates the geodesic curve $\cC_{\km,\s}$ and the red dot is the point $\s$. The green dot indicates the saddle point $\km$.  (\textbf{b}) The geodesic distance map $\cU_{\q}$. The black line denotes  $\cC_{\km,\q}$ and the yellow dot indicates $\q$.   (\textbf{c}) The white line denotes the concatenated  curve $\cC_{\s,\q}$. The blue shadow region is the constrained region $\bU$. (\textbf{d}) The radius-lifted geodesic curve $\hat\cC_{\hs,\hq}$. }
\label{fig:SaddlePointFP}
\end{figure*}

\noindent\emph{Metric visualization}.
We use  the control set, a tool for visualizing the anisotropy of a metric, to show the advantages of  the proposed metric $\cF_{\rm coh}$ at a crossing point. 
For the metric $\cF_{\rm coh}$ at a point $\x$, it can be defined by
\begin{equation}
\label{eq:ControlSet}
\cB_{\rm coh}(\x)=\{\fu\in\bR^2;~\cF_{\rm coh}(\x,\fu)\leq1\}.
\end{equation}
For comparison, we also consider an anisotropic metric 
\begin{equation*}
\tilde\cF_{\rm aniso}(\x,\fu)=\sqrt{\<\fu,\cM_{\rm aniso}(\x,\zeta_{\rm scale}(\x))\fu\>},
\end{equation*}
where  $\cM_{\rm aniso}$ is defined in Eq.~\eqref{eq:TensorDescirbeVessels}.  Let $\cB_{\rm aniso}$ be the control set of $\tilde\cF_{\rm aniso}$ constructed through Eq.~\eqref{eq:ControlSet}. For a Riemannian metric, the control set appears to be an ellipse  and we expect that  the direction of its major axis  should align with the vessel direction as much as possible. 

We construct the metrics  $\tilde\cF_{\rm aniso}$ and $\cF_{\rm coh}$ for the synthetic image shown  in Fig.~\ref{fig:anisotropyVector}a. We illustrate the control sets $\cB_{\rm aniso}(\x_{\rm min})$ and $\cB_{\rm coh}(\x_{\rm min})$ respectively  in Figs.~\ref{fig:ControlSet}a and \ref{fig:ControlSet}b, where the point $\x_{\rm min}$ is indicated by the green dot in Fig.~\ref{fig:ReferencePoints}b.  The red line in each figure  indicates the  target vessel direction at $\x_{\rm min}$ while each black line indicates the major axis of the corresponding ellipse.  One can point out that in Fig.~\ref{fig:ControlSet}a, the black and red lines are misaligned with each other since the tensor $\cM_{\rm aniso}(\x_{\rm min})$ is dominated by the strong vessel. While in Fig.~\ref{fig:ControlSet}b, the red and black lines are almost collinear to each other thanks to the adaptive anisotropic  tensor $T_{\rm aniso}(\x_{\rm min})$.

\begin{algorithm}[!t]
\caption{\textsc{Single~Front~Propagation~Scheme}}	
\label{algo:SingleFP}
\begin{algorithmic}
\renewcommand{\algorithmicrequire}{\textbf{Input:}}
\renewcommand{\algorithmicensure}{\textbf{Output:}}
\Require The orientation score $\Psi$, the points $\s$ and  $\q$.
\Ensure Geodesic distance map $\cU_\s$.
\renewcommand{\algorithmicrequire}{\textbf{Initialization:}}
\Require 
\State $\bullet$ Set $\cU_\s(\x)\gets\infty$ and $\cV(\x)\gets$ \emph{Far}, $\forall \x\in\Omega\backslash\{\s\}$.
\State $\bullet$ Set $\cU_\s(\s)\gets0, \cV(\s)\gets$ \emph{Front}, $\kp(\s)\gets\arg\max_{\theta}\Psi(\s,\theta)$.
\end{algorithmic}
\begin{algorithmic}[1]
\While{$\cV(\q)\neq$\emph{Accepted}} 
\State Find $\x_{\rm min}$, the \emph{Front}~point which minimizes $\cU_\s$.
\State Set $\cV(\x_{\rm min})\gets$ \emph{Accepted}.
\State Find the reference points $\a$ and $\b$ through Eq.~\eqref{eq:ReferencePoints}.
\For{All $\x_{\rm n}\in \cS^{-1}(\x_{\rm min})$ and $\cV(\x_{\rm n})\neq $ \emph{Accepted}}
\State Set $\cV(\x_{\rm n})\gets$ \emph{Front}.
\State Update $\kp(\x_{\rm n})$  by the reference point $\a$ via Eq.~\eqref{eq:AdmissOriens}.\label{line:dyn1}
\State Update $T_{\rm coh}(\x_{\rm n})$ by the point $\b$ via Eq.~\eqref{eq:DynamicTensor}.\label{line:dyn2}
\State Estimate $\cU_{\rm tem}(\x_{\rm n})$ by evaluating the Hopf-Lax operator in Eq.~\eqref{eq:HopfLax}.
\State Set $\cU_\s(\x_{\rm n}) \gets \min\{\cU_{\rm tem}(\x_{\rm n}),\cU_\s(\x_{\rm n})\}$.
\EndFor
\EndWhile
\end{algorithmic}
\end{algorithm}

\subsection{Partial Fronts Propagation Implementation}
\label{subsec:DualFP}
In the single front propagation scheme presented in Section~\ref{subsec:SinglePF}, in each geodesic distance update iteration, the back-tracked geodesic path $\bar{\cC}_{\x_{\rm min}}$ obtained from Eq.~\eqref{eq:physicalODE} always links the latest \emph{accepted} point $\x_{\rm min}$  to the source point $\s$. When the fast marching front arrives closely to the end point $\q$, we expect that the associated reference points are located at the vessel segment between  $\x_{\rm min}$ and $\q$ in order to obtain more adequate appearance coherence penalty. For this purpose, we consider the partial fronts propagation method~\cite{deschamps2001fast}. 

We can estimate the respective  geodesic distance maps $\cU_\s$ and $\cU_\q$ with  $\cU_\s(\s)=\cU_\q(\q)=0$ through the fast marching method with dynamic metric update scheme as presented in Algorithm~\ref{algo:SingleFP}. A saddle point $\km$ is the point which has the minimal value of $\cU_\s$ (or $\cU_\q$) among the equivalence  distance point set $\rA=\{\x;\,\cU_\s(\x)=\cU_\q(\x)\}$, i.e.,
\begin{equation}
\label{eq:Saddle}
\km=\mathop{\arg\,\!\min}_{\x\in\rA}\,\cU_\s(\x).
\end{equation}
We can track two geodesic curves $\bar{\cC}_{\km,\s}$ and $\bar{\cC}_{\km,\q}$ from the saddle point $\km$ through the solutions to  the gradient descent ODEs respectively on $\cU_{\s}$  and  $\cU_{\q}$. Let $\cC_{\s,\km},\,\cC_{\q,\km}\in\kL([0,1],\Omega)$ be the reversed and  re-parameterized curves of $\bar\cC_{\km,\s}$ and $\bar\cC_{\km,\q}$ respectively. The final geodesic curve $\cC_{\s,\q}$ with $\cC_{\s,\q}(0)=\s$ and $\cC_{\s,\q}(1)=\q$ can be obtained by concatenating the geodesic curves $\cC_{\km,\s}$ and $\cC_{\km,\q}$ as follows:
\begin{equation}
\label{eq:concatenation}	
\cC_{\s,\q}(v)=
\begin{cases}
\cC_{\s,\km}(2v),\quad&\text{if~} 0\leq v\leq1/2,\\
\cC_{\q,\km}(2(1-v)),\quad&\text{if~} 1/2< v\leq1.
\end{cases}
\end{equation}

For numerical implementation, we perform the fast marching front propagation as presented in Algorithm~\ref{algo:SingleFP} simultaneously from the  points $\s$ and $\q$. In this case, the saddle point $\km$ is the \emph{first} meeting point of the two fronts respectively expanding from  $\s$ and $\q$. That partial fronts propagation will be terminated once the saddle point $\km$ is detected in order to reduce the computation complexity. We illustrate an example for this partial fronts propagation scheme in Fig.~\ref{fig:SaddlePointFP}. In Figs.~\ref{fig:SaddlePointFP}a and \ref{fig:SaddlePointFP}b, the geodesic distance maps $\cU_\s$ and $\cU_\q$ together with the corresponding geodesic paths $\bar\cC_{\km,\s}$ and $\bar\cC_{\km,\q}$ are demonstrated.  The white line in Fig.~\ref{fig:SaddlePointFP}c indicates the concatenated curve  $\cC_{\s,\q}$.

\subsection{Region-Constrained Radius-lifted Geodesic Model}
\label{subsec:PriorConstrainedRegion}
The geodesic curves associated to the metric $\cF_{\rm coh}$ are computed in the image domain $\Omega$. However, for a complete tubular structure segmentation, the goal is to search for the centerline  and the corresponding thickness of the vessel simultaneously. Moreover,  we observe that the geodesic curves from $\cF_{\rm coh}$ sometimes suffer from a centerline bias problem, mainly because of the inhomogeneous intensity distributions. To solve these problems, we propose a region-constrained minimal path method, providing that a prescribed curve is given in order to build the constrained region. 

We take the  concatenated curve $\cC_{\s,\q}$ as an example of the prior curve.  Let $\bU\subset\Omega$ be a bounded and connected tubular neighbourhood of $\cC_{\s,\q}$ (see Fig.~\ref{fig:SaddlePointFP}c for an example of $\bU$), which can be efficiently computed by the morphological dilation operator. The region-constrained metric $\cF_{\rm cstr}:\hat\Omega\to[0,\infty]$ can be expressed for any point $\hx=(\x,r)$ by
\begin{equation}
\label{eq:RCMetric}
\cF_{\rm cstr}(\hx,\hat\fu)=
\begin{cases}
\|\hat\fu\|_{\cM_{\rm scale}(\hx)},&\forall \x\in\bU,\\
\infty,	&\text{otherwise},
\end{cases}	
\end{equation}
where $\cM_{\rm scale}$ is defined in Eq.~\eqref{eq:RadiusLiftedTensor}. The geodesic distance map associated  to $\cF_{\rm cstr}$ can be efficiently solved by the general anisotropic variant of the fast marching method~\cite{mirebeau2014anisotropic}, where the  propagation is terminated once the end point $\hat\q=(\q,\zeta_{\rm scale}(\q))$ is tagged as \emph{Accepted}. Obviously, the radius-lifted geodesic curve $\hat\cC_{\hs,\hq}=(\eta,\tau)$ associated to $\cF_{\rm cstr}$ satisfies $\eta(u)\in\bU,\,\forall u\in[0,1]$. We show the  geodesic curve $\hat\cC_{\hs,\hq}$ in Fig.~\ref{fig:SaddlePointFP}d, where the red curve denotes the centerline $\eta$ and the yellow contour depicts the vessel boundary derived from $\tau$. 

The region-constrained minimal path model can  seek a complete vessel segmentation in conjunction with the appearance feature coherence-penalized  minimal path model. In the following experiments, we take the paths  from the metric  $\cF_{\rm coh}$ as the prior curves to establish the respective tubular regions for the metric $\cF_{\rm cstr}$. 

\noindent \emph{Remark}.
The region-constrained minimal path model can be taken as an efficient way to estimate the vessel thickness measurement  from a  binary vessel segmentation map\footnote{Each grid point in this map is classified as either a vessel point or a background point.}. An interesting example as introduced in~\cite{chen2015piecewise} is to generate a  set of disjoint skeletons from the binary segmentation map. As a result, each skeleton can provide two end points and a tubular region for $\cF_{\rm cstr}$, yielding the thickness measurement for each vessel segment.

\section{Experimental Results}
\label{sec:Experiments}

\subsection{Parameter Setting}
\label{subsec:Parameters}
The orientation score $\Psi$ is computed by the oriented Gaussian kernels defined in Eq.~\eqref{eq:AsyOrientedGuassian}. In numerical implementations,  the variances $\sigma_1,\,\sigma_2$ of the oriented Gaussian kernels dominate the  anisotropy properties of these kernels. In the experiments, we fix $\sigma_1=300$ and $\sigma_2=1$  to construct  a series of well-oriented Gaussian kernels.  The parameter $w$ which controls the size of the truncated window for each oriented Gaussian kernel should depend on the image data. For instance,  if the target tubular structure  has strong tortuosity, the values of $w$ should be small. In case the target vessel crosses a stronger and thicker one, especially when the target is invisible at the crossing region,  a large value of $w$ is preferred. In our experiments, we  set $w=11$ unless specified otherwise.
The thresholding lengths  $\chi_1$ and $\chi_2$ (in grid point) of the back-tracked short geodesic curves are used to seek the two  reference points. The parameter $\chi_1$ contributes to the estimation of the vector field $\kp$ based on the Eqs.~\eqref{eq:optimalOrienCandidates} and \eqref{eq:AdmissOriens}. When the target has strong  tortuosity, the reference point associated to $\chi_1$ should be close to the latest \emph{Accepted} point,  which corresponds to a small $\chi_1$. The values of $\chi_2$ affect the appearance feature coherence penalty, which should be set dependently to the image data. In default, we experimentally set $\chi_1=1$ and $\chi_2=12$.  The parameters $\alpha$ in Eq.~\eqref{eq:StaticTensor2} and $\lambda$ in Eq.~\eqref{eq:SimilarityOS} control the influence from the tubular appearance and from the coherence penalty, respectively. For the case that a weak tubular structure crosses a strong one, the values of $\alpha$ should be low while the values of $\lambda$ should be high.  Finally, the constants  $\xi_{\rm ident}$ and $\xi_{\rm aniso}$ used in Eqs.~\eqref{eq:OrienScoreTensor} and \eqref{eq:anisoAdaptive} dominates the anisotropy property of the tensor fields $T_{\rm base}$ and $T_{\rm aniso}$. We declare the default values for these parameters as follows:  $\alpha=2$, $\lambda=20$, $\xi_{\rm aniso}=10$ and $\xi_{\rm ident}=0.1$. In the following experiments, we make use of the default setting discussed above unless specified otherwise. The experiments are performed on a standard Intel Core i$7$ of $4.2$GHz architecture with $32$Gb RAM. 

\begin{figure}
\centering
\includegraphics[width=4.2cm]{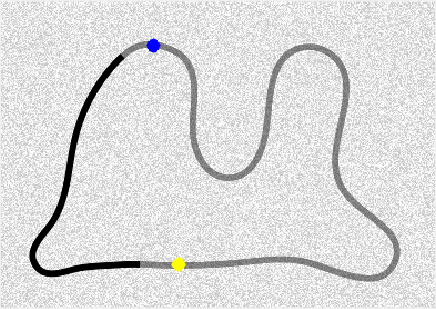}~\includegraphics[width=4.2cm]{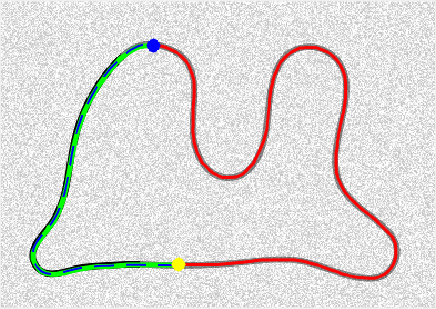}
\caption{Comparative results on a synthetic image. \textbf{Left} Prescribed points. \textbf{Right} Geodesic paths obtained from the metrics $\cF_{\rm aniso}$ (blue dash line), $\cF_{\rm e}$ (green solid line) and $\cF_{\rm coh}$ (red line). The objective is to extract the weak tubular structure.}
\label{fig:Synthetic}
\end{figure}

\begin{figure*}[t]
\centering
\includegraphics[width=3.2cm]{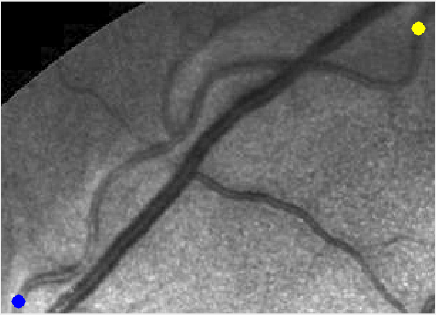}~\includegraphics[width=3.2cm]{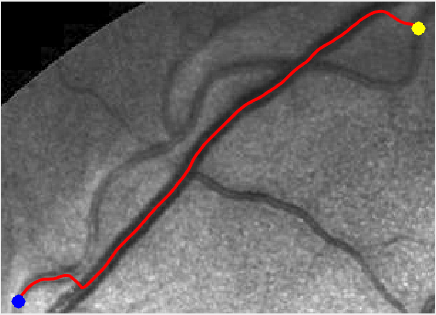}~\includegraphics[width=3.2cm]{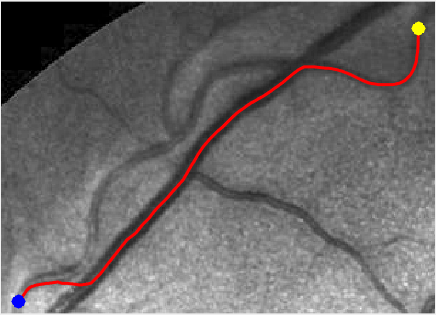}~\includegraphics[width=3.2cm]{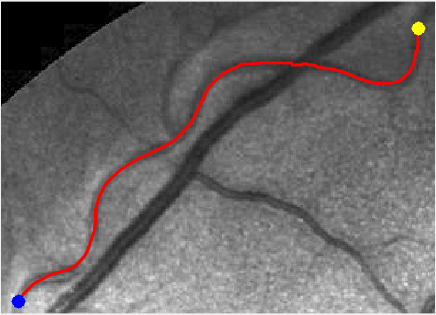}~\includegraphics[width=3.2cm]{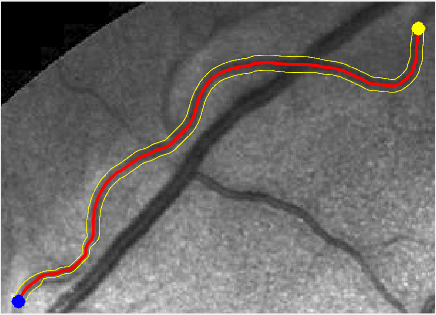}\\
\includegraphics[width=3.2cm]{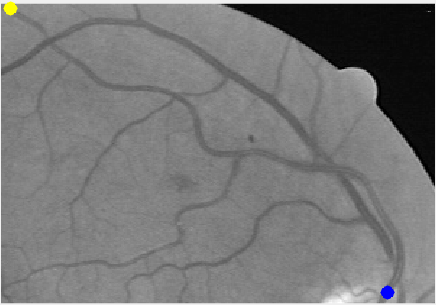}~\includegraphics[width=3.2cm]{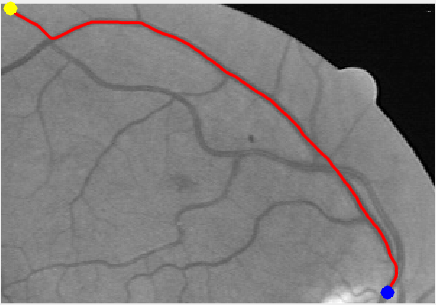}~\includegraphics[width=3.2cm]{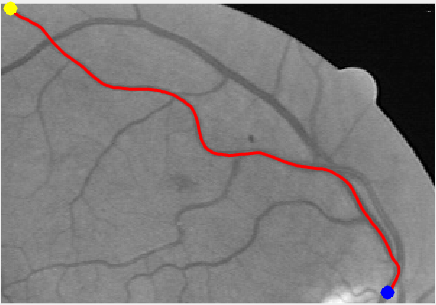}~\includegraphics[width=3.2cm]{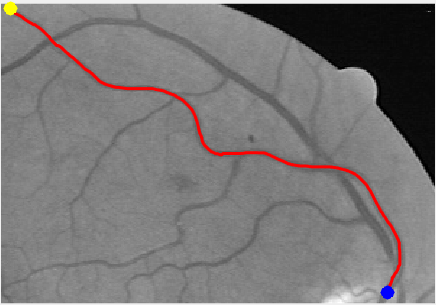}~\includegraphics[width=3.2cm]{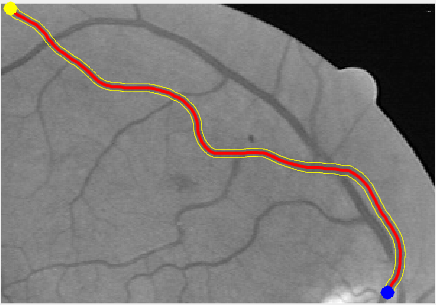}\\
\includegraphics[width=3.2cm]{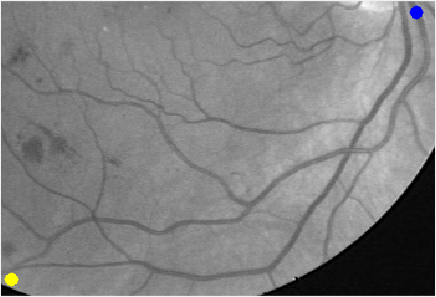}~\includegraphics[width=3.2cm]{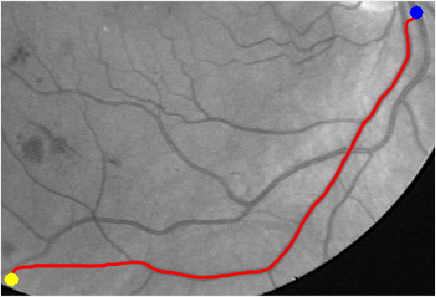}~\includegraphics[width=3.2cm]{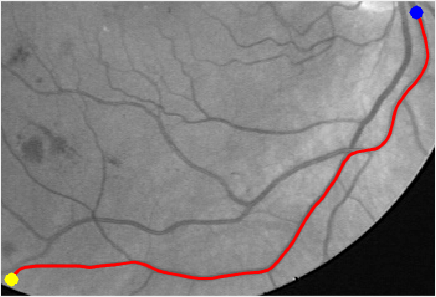}~\includegraphics[width=3.2cm]{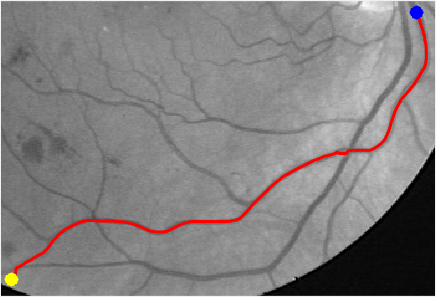}~\includegraphics[width=3.2cm]{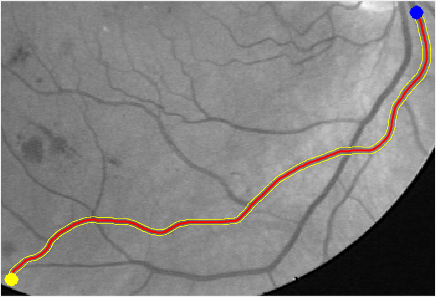}\\
\includegraphics[width=3.2cm]{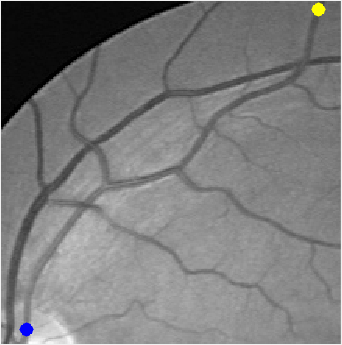}~\includegraphics[width=3.2cm]{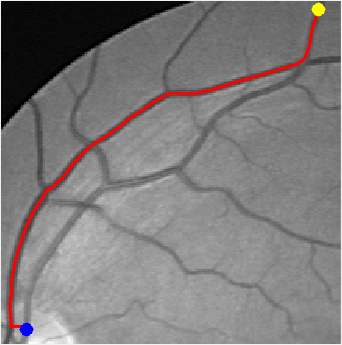}~\includegraphics[width=3.2cm]{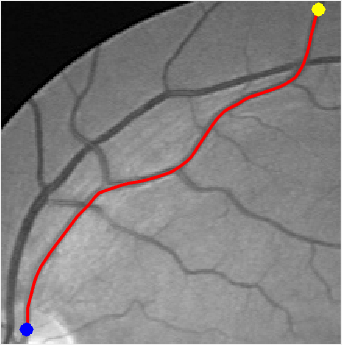}~\includegraphics[width=3.2cm]{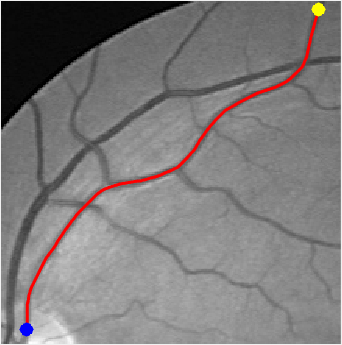}~\includegraphics[width=3.2cm]{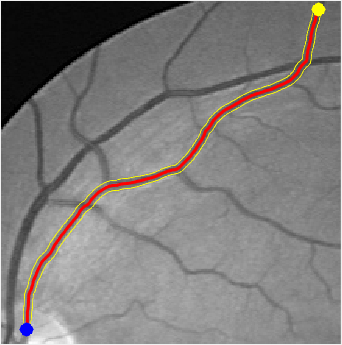}
\caption{Comparative results on retinal images. \textbf{Column 1} Prescribed points which are indicated by dots. \textbf{Columns 2-4} Minimal paths from the metrics of $\cF_{\rm aniso}$, $\cF_{\rm e}$,  the proposed $\cF_{\rm coh}$ and $\cF_{\rm cstr}$, respectively. Note that we only show the centerline positions of the minimal paths from the RLAR metric $\cF_{\rm aniso}$.}
\label{fig:DynExamples}
\end{figure*}

\begin{figure*}[t]
\centering
\includegraphics[width=4cm]{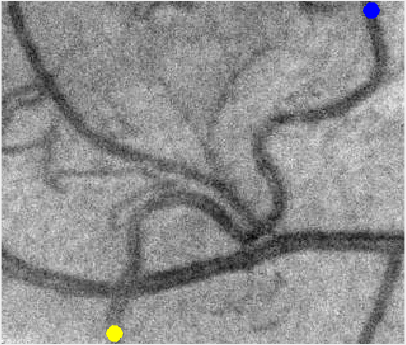}~\includegraphics[width=4cm]{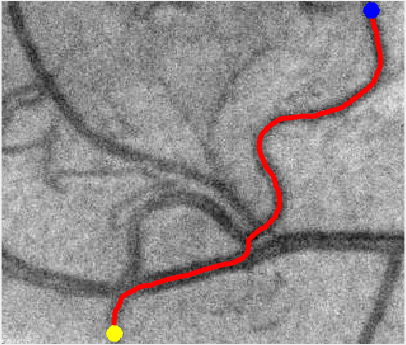}~\includegraphics[width=4cm]{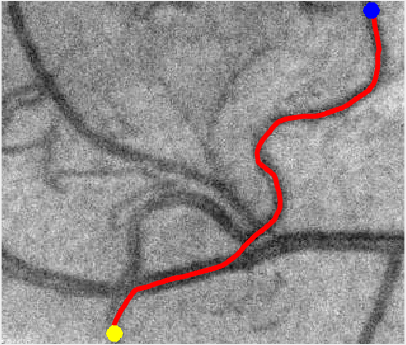}~\includegraphics[width=4cm]{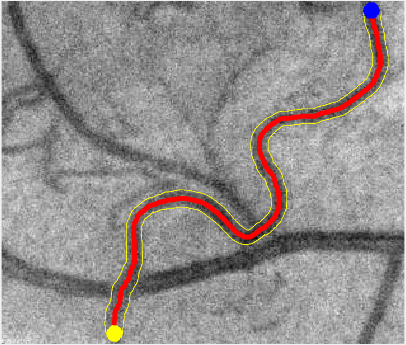}	\\
\includegraphics[width=4cm]{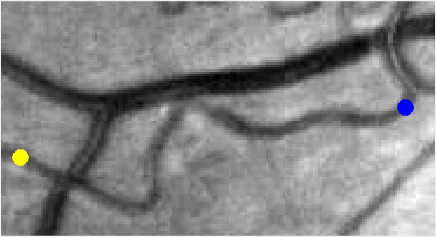}~\includegraphics[width=4cm]{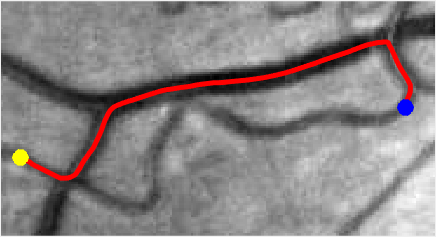}~\includegraphics[width=4cm]{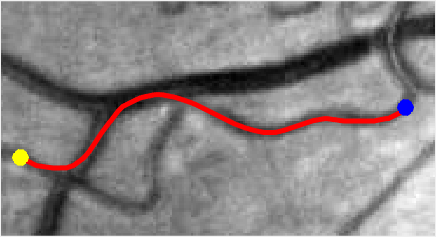}~\includegraphics[width=4cm]{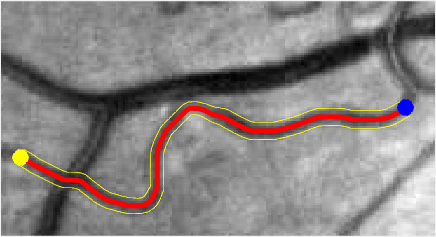}
\caption{Comparative results on retinal images with strong tortuosity. \textbf{Column 1} Prescribed points indicated by dots. \textbf{Columns 2-4} Minimal paths from the RLAR metric $\cF_{\rm aniso}$, the FE metric $\cF_{\rm e}$ and the proposed AFC metric $\cF_{\rm coh}$ respectively.}
\label{fig:Tortuosity}
\end{figure*}

\begin{figure}[t]
\centering
\includegraphics[width=2.0cm]{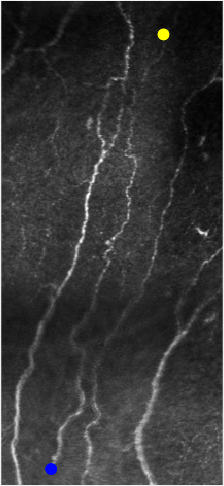}~\includegraphics[width=2.0cm]{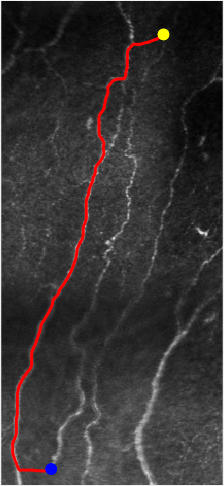}~\includegraphics[width=2.0cm]{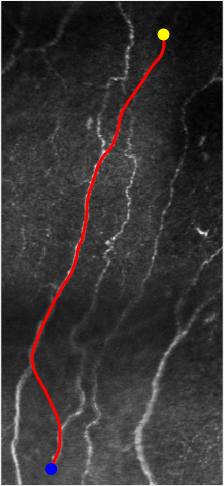}~\includegraphics[width=2.0cm]{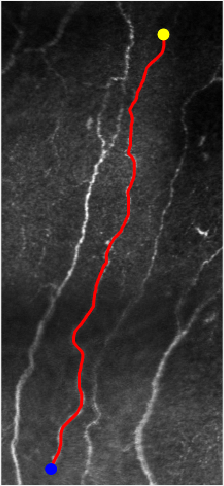}\\
\includegraphics[width=2.0cm]{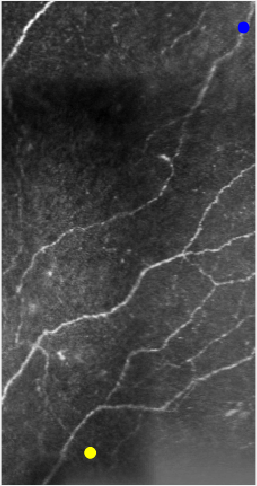}~\includegraphics[width=2.0cm]{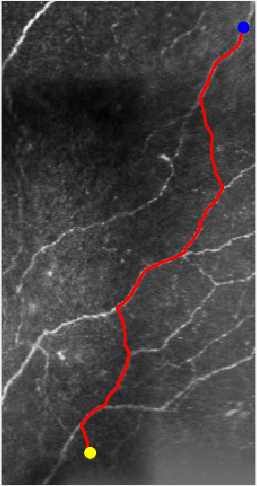}~\includegraphics[width=2.0cm]{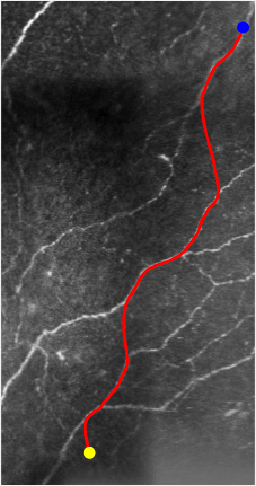}~\includegraphics[width=2.0cm]{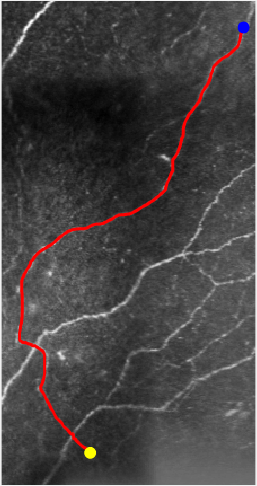}
\caption{Comparative results on neural fibre images. \textbf{Column 1} Prescribed source and end points indicated by blue and yellow dots. \textbf{Columns 2-4} Minimal paths (indicated by red lines) derived from the metrics of $\cF_{\rm aniso}$, $\cF_{\rm e}$ and $\cF_{\rm coh}$, respectively.}
\label{fig:Fiber}
\end{figure}

\begin{figure}[t]
\centering
\subfigure[]{\includegraphics[width=4.2cm,angle=90]{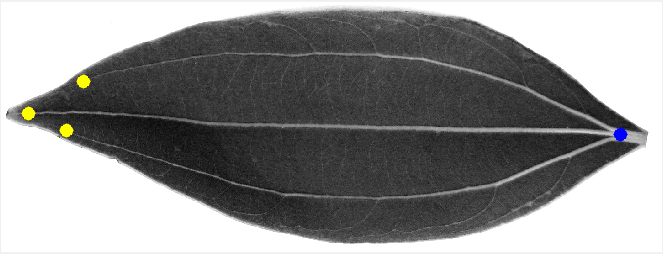}}~\subfigure[]{\includegraphics[width=4.2cm,angle=90]{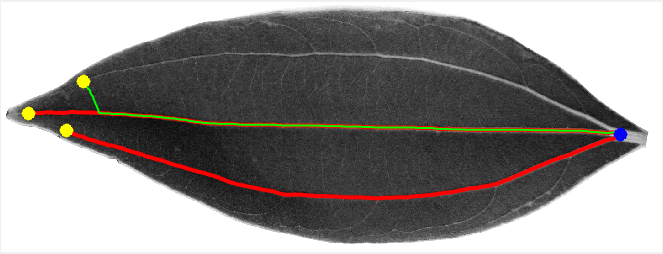}}~\subfigure[]{\includegraphics[width=4.2cm,angle=90]{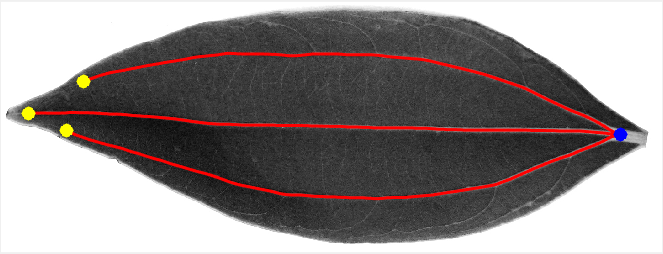}}~\subfigure[]{\includegraphics[width=4.2cm,angle=90]{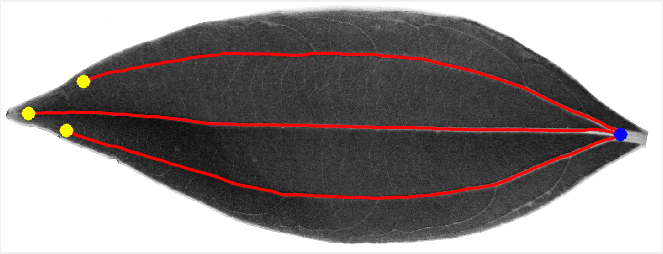}}~\subfigure[]{\includegraphics[width=4.2cm,angle=90]{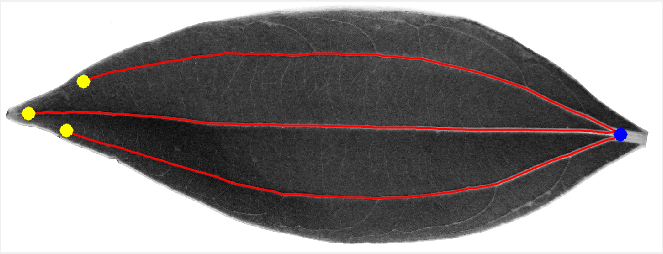}}
\caption{Comparative results on a leaf image. \textbf{(a)} The leaf image with source point (blue dots) and end points (yellow dots). \textbf{(b)}-\textbf{(e)}: Geodesic curves from the metrics $\cF_{\rm aniso}$, $\cF_{\rm e}$, $\cF_{\rm coh}$ and $\cF_{\rm cstr}$ respectively. }
\label{fig:Leaf}
\end{figure}

\subsection{Comparative Results with State-of-the-art Metrics}
We compare the proposed metrics, including the appearance feature coherence penalized (AFC) metric  $\cF_{\rm coh}$ and the region-constrained (RC) metric $\cF_{\rm cstr}$, to the radius-lifted anisotropic  Riemannian (RLAR) metric~\cite{benmansour2011tubular} and the Finsler elastica (FE)  metric~\cite{chen2017global}. By the tensor field $\cM_{\rm scale}$ in Eq.~\eqref{eq:RadiusLiftedTensor}, the RLAR metric $\cF_{\rm aniso}:\hat\Omega\to[0,\infty]$ can be formulated by
\begin{equation}
\label{eq:RLARMetric}
\cF_{\rm aniso}(\hx,\hat\fu)=\|\hat\fu\|_{\cM_{\rm scale}(\hx)}.
\end{equation}
The anisotropy ratio~\cite{benmansour2011tubular} for $\cF_{\rm aniso}$ can be defined by
\begin{equation*}
C_{\rm ratio}(\cF_{\rm aniso})=\max_{\hx\in\Omega}\left\{\max_{\|\fu_1\|=\|\fu_2\|=1}\frac{\|\fu_1\|_{\cM_{\rm aniso}(\hx)}}{\|\fu_2\|_{\cM_{\rm aniso}(\hx)}}\right\},
\end{equation*}
where $\cM_{\rm aniso}$ is defined in Eq.~\eqref{eq:TensorDescirbeVessels}. The values of the parameter $\alpha_{\rm aniso}$ (see Eq.~\eqref{eq:TensorDescirbeVessels}) can be determined by  the ratio $C_{\rm ratio}$. We set $C_{\rm ratio}=10$ in the following experiments.  The computation of the geodesic distance associated with $\cF_{\rm aniso}$ will be terminated once all the end points have been reached. 

The FE metric $F_{\rm e}:\Omega\times\bS^1\to[0,\infty]$ is established in an orientation-lifted space~\cite{chen2017global},  which can be expressed by
\begin{equation*}
\cF_{\rm e}(\tilde\x,\tilde\fu)=P_{\rm os}(\tilde\x)(\sqrt{\rho^2\|\fu\|^2+2\rho\beta_{\rm ela} |\nu|^2}-(\rho-1)\<\kg(\theta),\fu\>)
\end{equation*}
where $\tilde\x=(\x,\theta)\in\Omega\times\bS^1$ is an orientation-lifted point and $\tilde\fu=(\fu, \nu)\in\bR^3$ is a vector. The value of the parameter $\rho$ determining the anisotropy and asymmetry  properties of the metric $\cF_{\rm e}$ is fixed to $100$ for all the experiments.  The function $P_{\rm os}:\Omega\times\bS^1\to\bR^+$ can be derived from  the orientation score $\psi$ defined in Eq.~\eqref{eq:OS}, i.e., 
\begin{equation*}
P_{\rm os}(\x,\theta)=\exp(-\alpha_{\rm ela}\,\psi(\x,\theta)/\|\psi\|_\infty),\quad \alpha_{\rm ela}\in\bR^+.
\end{equation*}
The values of $\alpha_{\rm ela}$ and $\beta_{\rm ela}\in\bR^+$ control the balance between the curvature penalization and the image data~\cite{chen2017global}, which should be tuned dependently on the target structures.  In our  experiments, we adopt the same Eikonal solver as used in~\cite{chen2017global}, i.e., the Finsler variant of the fast marching method~\cite{mirebeau2014efficient}, for the FE metric $\cF_{\rm e}$. We also use the same tubular structure extraction strategy  as proposed in~\cite{chen2017global}.  

In Fig.~\ref{fig:Synthetic}, we compare the proposed AFC metric $\cF_{\rm coh}$ with the RLAR metric $\cF_{\rm aniso}$ and the FE metric $\cF_{\rm e}$. The  structure in  Fig.~\ref{fig:Synthetic} is comprised of a strong tubular segment and a weak one. The goal is to extract the weak structure between two points. Both the minimal paths from $\cF_{\rm aniso}$ (blue dash line) and $\cF_{\rm e}$ (green solid line) prefer to pass through the tubular segment with strong appearance features. In contrast,  the geodesic curve (red line) associated to $\cF_{\rm coh}$ can delineate the desired structure. In this experiment, we use the default parameters for $\cF_{\rm coh}$  except for  $\chi_2$ which is set to $15$.

In Fig.~\ref{fig:DynExamples}, the minimal paths derived from the RLAR metric $\cF_{\rm aniso}$, the FE metric $\cF_{\rm e}$, the AFC metric $\cF_{\rm coh}$ and the RC metric $\cF_{\rm cstr}$ are shown in columns $2$ to $5$  respectively. In the first column, the prescribed  points are indicated by the green and yellow dots. In column $2$, the geodesic paths yielded by the RLAR metric $\cF_{\rm aniso}$  suffered from the  short branches combination problem, where these paths prefer to pass through the vessel segments with strong appearance features. In the first three rows of column $3$, we also observe the short branches combination problem  for the geodesic paths from the FE metric $\cF_{\rm e}$. In column $4$, the geodesic paths from the AFC metric $\cF_{\rm coh}$ can correctly depict the desired vessel segments due to the appearance feature coherence penalization. In column $5$,  we illustrate the radius-lifted minimal paths where the yellow contours  delineate the vessel boundaries and the red curves indicate the vessel centerlines. One can claim that the geodesic paths in  column $5$ is capable of accurately describing the target vessels. In rows $1$ to $3$, we use the default parameters for the metric $\cF_{\rm coh}$. While in row $4$, we  set $\lambda=10$.

In some extent, the short branches combination problem can be solved by the FE metric $\cF_{\rm e}$, as shown in column $3$ and row $4$ of Fig.~\ref{fig:DynExamples}. However, it is difficult for the FE metric to get the accurate results in the situation of extracting a weak vessel with strong tortuosity especially  when the target is close to   another vessel with strong appearance features. We show two such  examples in Fig.~\ref{fig:Tortuosity}. In columns $2$ to $4$ of Fig.~\ref{fig:Tortuosity}, the results from the RLAR metric $\cF_{\rm aniso}$, the FE metric $\cF_{\rm e}$ and the RC metric $\cF_{\rm cstr}$ are shown, indicating that only the proposed metric $\cF_{\rm cstr}$  can get the expected paths. 

In Fig.~\ref{fig:Fiber} we show the results from the RLAR metric $\cF_{\rm aniso}$, the FE metric $\cF_{\rm e}$ and the AFC metric $\cF_{\rm coh}$ on two neural fibre images\footnote{Many thanks to Dr. Tos T. J. M. Berendschot  from Maastricht University for providing the data.}. The curvilinear structures are treated as thin vessels. Some portions of the targets between the blue and yellow dots are weakly defined. Both the metrics $\cF_{\rm aniso}$ and $\cF_{\rm e}$ fail to detect the expected structures, while the minimal paths from the metric $\cF_{\rm coh}$ are able to accurately depict the targets. In this experiment, we use the default parameters for the metric $\cF_{\rm coh}$ except for  $\chi_2$ which is set to $5$.
The parameters $\alpha_{\rm ela}$ and $\beta_{\rm ela}$ for the FE metric $\cF_{\rm e}$ are set to $3$ and $50$  respectively.

In Fig.~\ref{fig:Leaf} we show the geodesic paths on a  leaf image~\cite{wu2007leaf} with respect to the metrics $\cF_{\rm aniso}$, $\cF_{\rm e}$, $\cF_{\rm coh}$ and $\cF_{\rm cstr}$. The geodesic paths shown in Fig.~\ref{fig:Leaf}b from the RLAR metric $\cF_{\rm aniso}$  fail to detect the left vein. In Figs.~\ref{fig:Leaf}c, \ref{fig:Leaf}d and~\ref{fig:Leaf}e, the geodesic paths derived from the metrics $\cF_{\rm e}$, $\cF_{\rm coh}$ and $\cF_{\rm cstr}$ are able to obtain the desired results. However, the computation time associated to the metric $\cF_{\rm e}$ in Fig~\ref{fig:Leaf}c requires about $77$ seconds for the leaf image of size $548\times1447$, while the computation time in Fig~\ref{fig:Leaf}d is only around $8$ seconds involving the construction of $T_{\rm base}$ in Eq.~\eqref{eq:StaticTensor2} and the geodesic distance computation. Note that for the results in Fig.~\ref{fig:Leaf}e, we only show the centerline positions denoted by the first two coordinates in each $3$D point in the obtained geodesic path. In this experiment, we use the single front propagation scheme for the results in Fig.~\ref{fig:Leaf}d.

\begin{table}[!t]
\centering
\caption{Quantitative comparisons of  different metrics on DRIVE.}
\label{tab:Result-DRIVE}
\setlength{\tabcolsep}{5pt}
\renewcommand{\arraystretch}{1.1}
\begin{tabular}{c||l c c c c}
\shline
\multicolumn{2}{c}{$\mathbb A$}$~\quad$  & $\cF_{\rm aniso}$ &$\cF_{\rm e}$  & $\cF_{\rm coh}$ & $\cF_{\rm cstr}$\\ 
\hline
\multirow{4}{*}{Artery Region}   & Avg.  &0.36 & 0.65 &0.92 &\textbf{0.98} \\
                                   & Max.  &1    & 1    & 1   &1    \\
                                   & Min.  &0.02 & 0.13 &0.60 &0.79     \\
                                   & Std.  &0.26 & 0.29 &0.08 &0.04      \\
\hline   
\multirow{4}{*}{Dilated Skeleton}    & Avg.  & 0.32 &0.53 &0.76  &\textbf{0.90}    \\
                                     & Max.  & 0.95 &0.94 &0.93  &0.99    \\
                                     & Min.  & 0.02 &0.12 &0.35  &0.5    \\
                                     & Std.  & 0.25 &0.25 &0.12  &0.08    \\
\shline
\end{tabular}
\end{table}

\begin{table}[!t]
\centering
\caption{Quantitative comparisons of  different metrics on IOSTAR.}
\label{tab:Result-IOSTAR}
\setlength{\tabcolsep}{6pt}
\renewcommand{\arraystretch}{1.1}
\begin{tabular}{l||l c c c c }
\shline
\multicolumn{2}{c}{$\mathbb A$}$~\quad$  & $\cF_{\rm aniso}$ &$\cF_{\rm e}$ & $\cF_{\rm coh}$ & $\cF_{\rm cstr}$ \\ 
\hline
\multirow{4}{*}{Dilated Artery Region}   & Avg.  &0.52 & 0.78 &0.93 &\textbf{0.95}     \\
                               & Max.  &0.99 & 1    &1    &1     \\
                               & Min.  &0.03 & 0.03 &0.53 &0.54     \\
                               & Std.  &0.34 & 0.33 &0.08 &0.08     \\
\shline
\end{tabular}
\end{table}

\subsection{Quantitative Comparative Results}
In this section, we quantitatively compare  the vessel detection  performance of  the proposed metrics $\cF_{\rm coh}$ and $\cF_{\rm cstr}$ with  $\cF_{\rm aniso}$ and $\cF_{\rm e}$ on  $88$  patches of retinal images from the DRIVE and IOSTAR datasets\footnote{We derive $45$ patches from the DRIVE dataset~\cite{staal2004ridge} and  $43$ patches from the IOSTAR dataset~\cite{zhang2016robust}.}. The  corresponding artery-vein (A-V) labeled images of the DRIVE and IOSTAR datasets are provided by~\cite{hu2013automated} and \cite{zhang2016robust}, respectively. Each patch includes a retinal artery which is close to or crosses  a vein with stronger appearance features. The objective is to extract the artery centerline between two given points. 

Let $\mathbb{A}$ be the set of the grid points inside the target artery region derived from the respective A-V label image and let $|\mathbb{A}|$ be the number of elements of the set $\mathbb{A}$. In addition, we define a set $\Gamma$ of grid points passed through by a continuous geodesic curve. A scalar-valued measure $\Theta\in[0,1]$   can be  defined by $\Theta=|\Gamma\cap\mathbb{A}|/|\Gamma|$.

For the DRIVE dataset, we provide two ways to construct the set $\mathbb{A}$ from the A-V labeled images. The first way is to regard  $\mathbb A$ as the set of all the grid points tagged as artery.  The second way is to skeletonize  the regions labeled as artery via the morphological operators, followed by a dilation operation with radius $1$ on these skeletons. Then we remove the non-vessel grid points from the dilated skeletons. In Table.~\ref{tab:Result-DRIVE}, the two construction methods are referred to as \emph{Artery Region} and \emph{Dilated Skeleton} respectively. 
For the IOSTAR dataset, we directly dilate the regions tagged as artery  by a morphological operator with radius $1$, which is named \emph{Dilated Artery Region}. The use of the dilation operator is to mitigate the influences from the strong intensity inhomogeneities of the images in IOSTAR dataset. The results from $\cF_{\rm coh}$ shown  in Table~\ref{tab:Result-DRIVE} are obtained using the default  parameters described in Section~\ref{subsec:Parameters}. In Table~\ref{tab:Result-IOSTAR},  $\alpha$ and $\lambda$ for the metric $\cF_{\rm coh}$ are chosen from the sets $\{1,2,3\}$ and $\{10,20,30\}$, respectively  to mitigate the effects from the retinal vessel centre reflection. The quantitively comparative  results on the  DRIVE and  IOSTAR datasets associated to the metrics $\cF_{\rm aniso}$, $\cF_{\rm e}$,  $\cF_{\rm coh}$ and $\cF_{\rm cstr}$  are respectively shown in Tables.~\ref{tab:Result-DRIVE} and~\ref{tab:Result-IOSTAR}. 

One can claim that the AFC metric $\cF_{\rm coh}$ and the RC metric $\cF_{\rm cstr}$ outperform the state-of-the-art metrics $\cF_{\rm aniso}$ and $\cF_{\rm e}$  on both datasets. The results from $\cF_{\rm aniso}$ correspond to the lowest values of $\Theta$ in both datasets due to  the short branches combination and shortcut problems. The elastica geodesic paths try to avoid sharp turnings as much as possible due to the curvature penalization. This property  matches the observation of the retinal arteries,  leading to a better performance than the metric $\cF_{\rm cstr}$. However, sometimes the FE metric still suffers from the short branches combination and shortcut problems due to the weak appearance or  high tortuosity features of the targets. In addition, the average computation time (in seconds)  for the metrics $\cF_{\rm aniso}$, $\cF_{\rm e}$,  $\cF_{\rm coh}$ and  $\cF_{\rm cstr}$ are $0.58$s, $0.71$s, $0.35$s  and $0.11$s for the DRIVE retinal patches, respectively. For the IOSTAR dataset, the computation time are $3.85$s, $8.05$s,  $1.14$s and $0.65$s respectively. Note that the computation time for the  metric $\cF_{\rm coh}$ includes the construction of the sets $\kM_\x$, the tensor field $T_{\rm base}$ and the geodesic distance computation. 
The experimental results in Tables.~\ref{tab:Result-DRIVE} and~\ref{tab:Result-IOSTAR} show that the proposed metrics $\cF_{\rm coh}$ and $\cF_{\rm cstr}$ are indeed  effective for the retinal vessel extraction.

\begin{figure}
\centering
\includegraphics[height=3.1cm]{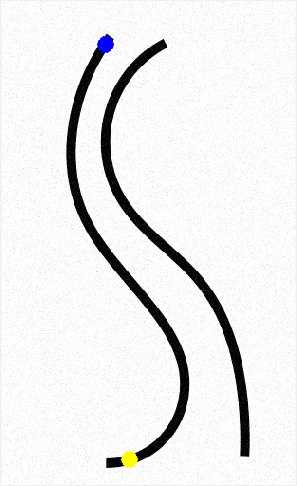}~\includegraphics[height=3.1cm]{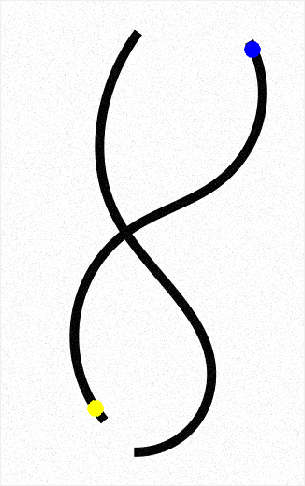}~\includegraphics[height=3.1cm]{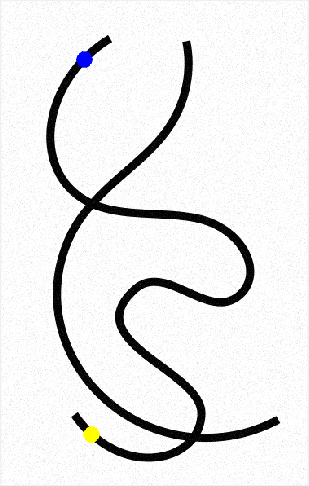}~\includegraphics[height=3.1cm]{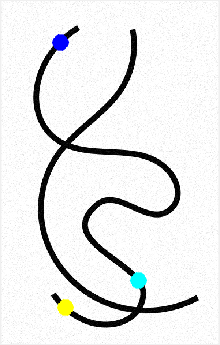}\\
\includegraphics[height=3.1cm]{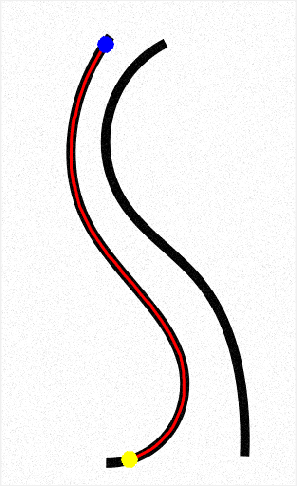}~\includegraphics[height=3.1cm]{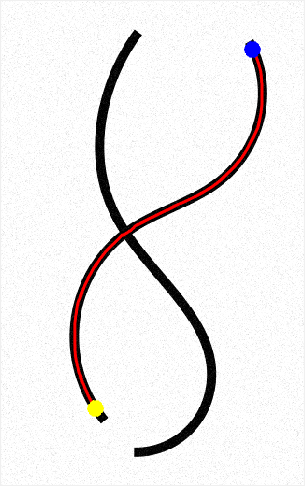}~\includegraphics[height=3.1cm]{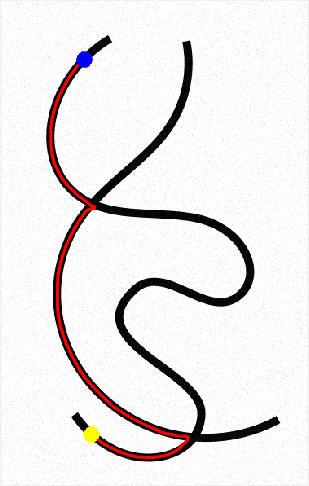}~\includegraphics[height=3.1cm]{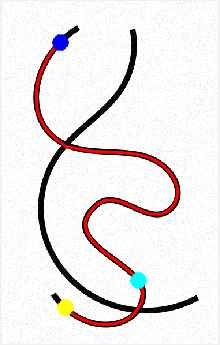}
\caption{The minimal paths on different synthetic images. \textbf{Row 1} The original images and the prescribed points. \textbf{Row 2} The corresponding minimal paths obtained using the metric $\cF_{\rm coh}$.}
\label{fig:Limits}
\end{figure}

In Fig.~\ref{fig:Limits} we show the geodesic paths derived from the AFC metric $\cF_{\rm coh}$ in the case that the gray levels of the targets are almost identical  to  its neighbouring structures. In  column $1$, the two structures are close to each other without overlapping, while in column $2$, a tubular  structure crosses another one just once.  In both cases, the metric $\cF_{\rm coh}$ are able to get the expected results. In column $3$, the two tubular structures yield a loop, where the target is longer than another one in the sense of Euclidean length. The geodesic path $\cC_{\q_1,\q_2}$ between the points $\q_1$ (blue dot) and $\q_2$ (yellow dot) from the metric $\cF_{\rm coh}$ fails to follow the target. In order to solve this problem, one can simply add a new point $\q_3$ (cyan dot) to the target at the loop region, as shown in column $4$ of Fig.~\ref{fig:Limits}. The geodesic paths $\cC_{\q_1,\q_3}$ and $\cC_{\q_2,\q_3}$ can be concatenated to form the final path.

\section{Conclusion}
\label{Sec:Conclusion}
In this paper, we propose two minimal path models: a dynamic model and a region-constrained model. The first model relies on a new metric which blends the benefits from  the appearance feature coherence penalization and the adaptive anisotropy enhancement. This metric is constructed dynamically  during  the front propagation. The dynamic minimal path model can correctly extract a vessel between two points from a complicated tree structure providing that along the target the appearance features vary smoothly. The second model is posed in a radius-lifted domain established by adding a radius dimension to a tubular  neighbourhood of a prescribed curve. The integration of the two  models can seek a complete segmentation of a vessel, and meanwhile can reduce the risk of the short branches combination and shortcut problems.

The future work for the proposed models will be dedicated to exploit more applications for tubular structure extraction  such as road detection in remote sensing images. In addition, we will also extend the proposed dynamic minimal path model for $2$D and $3$D vessel tree extraction applications in conjunction with the geodesic voting scheme. Such an extension is very natural and the initialization can be simplified to a single source point placed at the tree root.

\section*{Acknowledgment}
The authors thank the reviewers for their suggestions to improve the presentation of this paper. 
The authors also thank Dr.~Jean-Marie Mirebeau from Universit\'e Paris-Sud for his fruitful discussion. 
This research has been partially funded by Roche pharma (project AMD\_short) and by a grant from the French Agence Nationale de la Recherche ANR-16-RHUS-0004 (RHU TRT\_cSVD).

\ifCLASSOPTIONcaptionsoff
 \newpage
\fi

\bibliographystyle{IEEEbib}
\bibliography{minimalpath}

\begin{thebibliography}{10}

\bibitem{kirbas2004review}
C.~Kirbas and F.~Quek,
\newblock ``A review of vessel extraction techniques and algorithms,''
\newblock {\em ACM Comput. Surv.}, vol. 36, no. 2, pp. 81--121, 2004.

\bibitem{lesage2009review}
D.~Lesage, E.~D. Angelini, I.~Bloch, and G.~Funka-Lea,
\newblock ``{A review of 3D vessel lumen segmentation techniques: Models,
  features and extraction schemes},''
\newblock {\em Med. Image Anal.}, vol. 13, no. 6, pp. 819--845, 2009.

\bibitem{moccia2018blood}
S.~Moccia, E.~De~Momi, S.~El~Hadji, and L.~S. Mattos,
\newblock ``{Blood vessel segmentation algorithms—Review of methods, datasets
  and evaluation metrics},''
\newblock {\em Comput. Methods Programs Biomed.}, vol. 158, pp. 71--91, 2018.

\bibitem{lorigo2001curves}
L.~M. Lorigo et~al.,
\newblock ``Curves: Curve evolution for vessel segmentation,''
\newblock {\em Med. Image Anal.}, vol. 5, no. 3, pp. 195--206, 2001.

\bibitem{gooya2007effective}
A.~Gooya et~al.,
\newblock ``{Effective statistical edge integration using a flux maximizing
  scheme for volumetric vascular segmentation in MRA},''
\newblock in {\em Proc. IPMI}, 2007, pp. 86--97.

\bibitem{gooya2008r}
A.~Gooya, T.~Dohi, I.~Sakuma, and H.~Liao,
\newblock ``{R-PLUS: a Riemannian anisotropic edge detection scheme for
  vascular segmentation},''
\newblock in {\em Proc. MICCAI}, 2008, pp. 262--269.

\bibitem{peyre2010geodesic}
G.~Peyr{\'e}, M.~P{\'e}chaud, R.~Keriven, and L.~D. Cohen,
\newblock ``Geodesic methods in computer vision and graphics,''
\newblock {\em Foundations and Trends in Computer Graphics and Vision}, vol. 5,
  no. 3--4, pp. 197--397, 2010.

\bibitem{vasilevskiy2002flux}
A.~Vasilevskiy and K.~Siddiqi,
\newblock ``Flux maximizing geometric flows,''
\newblock {\em IEEE Trans. Pattern Anal. Mach. Intell.}, vol. 24, no. 12, pp.
  1565--1578, 2002.

\bibitem{law2009efficient}
M.~W. Law and A.~C. Chung,
\newblock ``Efficient implementation for spherical flux computation and its
  application to vascular segmentation,''
\newblock {\em IEEE Trans. Image Process.}, vol. 18, no. 3, pp. 596--612, 2009.

\bibitem{law2007weighted}
M.~W. Law and A.~C. Chung,
\newblock ``{Weighted local variance-based edge detection and its application
  to vascular segmentation in magnetic resonance angiography},''
\newblock {\em IEEE Trans. Med. Imag.}, vol. 26, no. 9, pp. 1224--1241, 2007.

\bibitem{descoteaux2008geometric}
M.~Descoteaux, D.~L. Collins, and K.~Siddiqi,
\newblock ``{A geometric flow for segmenting vasculature in proton-density
  weighted MRI},''
\newblock {\em Med. Image Anal.}, vol. 12, no. 4, pp. 497--513, 2008.

\bibitem{law2010oriented}
M.~W. Law and A.~C. Chung,
\newblock ``An oriented flux symmetry based active contour model for three
  dimensional vessel segmentation,''
\newblock in {\em Proc. ECCV}, 2010, pp. 720--734.

\bibitem{law2008three}
M.~W. Law and A.~C. Chung,
\newblock ``Three dimensional curvilinear structure detection using optimally
  oriented flux,''
\newblock in {\em Proc. ECCV}, 2008, pp. 368--382.

\bibitem{nain2004vessel}
D.~Nain, A.~Yezzi, and G.~Turk,
\newblock ``Vessel segmentation using a shape driven flow,''
\newblock in {\em Proc. MICCAI}, 2004, pp. 51--59.

\bibitem{gooya2008variational}
A.~Gooya et~al.,
\newblock ``A variational method for geometric regularization of vascular
  segmentation in medical images,''
\newblock {\em IEEE Trans. Image Process.}, vol. 17, no. 8, pp. 1295--1312,
  2008.

\bibitem{manniesing2007vessel}
R.~Manniesing, M.~A. Viergever, and W.~J. Niessen,
\newblock ``Vessel axis tracking using topology constrained surface
  evolution,''
\newblock {\em IEEE Trans. Med. Imag.}, vol. 26, no. 3, pp. 309--316, 2007.

\bibitem{cohen2007segmentation}
L.~D. Cohen and T.~Deschamps,
\newblock ``{Segmentation of 3D tubular objects with adaptive front propagation
  and minimal tree extraction for 3D medical imaging},''
\newblock {\em Comput. Methods Biomech. Biomed. Engin.}, vol. 10, no. 4, pp.
  289--305, 2007.

\bibitem{malladi1998real}
R.~Malladi and J.~A. Sethian,
\newblock ``A real-time algorithm for medical shape recovery,''
\newblock in {\em Proc. ICCV}, 1998, pp. 304--310.

\bibitem{chen2018fast}
D.~Chen and L.~D. Cohen,
\newblock ``Fast asymmetric fronts propagation for image segmentation,''
\newblock {\em J. Math. Imag. Vis.}, vol. 60, no. 6, pp. 766--783, 2018.

\bibitem{sato1998three}
Yoshinobu Sato et~al.,
\newblock ``Three-dimensional multi-scale line filter for segmentation and
  visualization of curvilinear structures in medical images,''
\newblock {\em Med. Image Anal.}, vol. 2, no. 2, pp. 143--168, 1998.

\bibitem{frangi1998multiscale}
A.~F. Frangi, W.~J. Niessen, K.~L. Vincken, and M.~A. Viergever,
\newblock ``Multiscale vessel enhancement filtering,''
\newblock in {\em Proc. MICCAI}, 1998, pp. 130--137.

\bibitem{jacob2004design}
M.~Jacob and M.~Unser,
\newblock ``Design of steerable filters for feature detection using canny-like
  criteria,''
\newblock {\em IEEE Trans. Pattern Anal. Mach. Intell.}, vol. 26, no. 8, pp.
  1007--1019, 2004.

\bibitem{moriconi2017vtrails}
S.~Moriconi et~al.,
\newblock ``Vtrails: Inferring vessels with geodesic connectivity trees,''
\newblock in {\em Proc. IPMI}, 2017, pp. 672--684.

\bibitem{franken2009crossing}
E.~Franken and R.~Duits,
\newblock ``Crossing-preserving coherence-enhancing diffusion on invertible
  orientation scores,''
\newblock {\em Int. J. Comput. Vis.}, vol. 85, no. 3, pp. 253, 2009.

\bibitem{hannink2014crossing}
J.~Hannink, R.~Duits, and E.~Bekkers,
\newblock ``Crossing-preserving multi-scale vesselness,''
\newblock in {\em Proc. MICCAI}, 2014, pp. 603--610.

\bibitem{zhang2016robust}
J.~Zhang et~al.,
\newblock ``Robust retinal vessel segmentation via locally adaptive derivative
  frames in orientation scores,''
\newblock {\em IEEE Trans. Med. Imag.}, vol. 35, no. 12, pp. 2631--2644, 2016.

\bibitem{merveille2018curvilinear}
O.~Merveille, H.~Talbot, L.~Najman, and N.~Passat,
\newblock ``Curvilinear structure analysis by ranking the orientation responses
  of path operators,''
\newblock {\em IEEE Trans. Pattern Anal. Mach. Intell.}, vol. 40, no. 2, pp.
  304--317, 2018.

\bibitem{poon2007live}
K.~Poon, G.~Hamarneh, and R.~Abugharbieh,
\newblock ``{Live-vessel: Extending livewire for simultaneous extraction of
  optimal medial and boundary paths in vascular images},''
\newblock in {\em Proc. MICCAI}, 2007, pp. 444--451.

\bibitem{ulen2015shortest}
J.~Ulen, P.~Strandmark, and F.~Kahl,
\newblock ``{Shortest paths with higher-order regularization},''
\newblock {\em IEEE Trans. Pattern Anal. Mach. Intell.}, vol. 37, no. 12, pp.
  2588--2600, 2015.

\bibitem{cohen1997global}
L.~D. Cohen and R.~Kimmel,
\newblock ``{Global minimum for active contour models:~A minimal path
  approach},''
\newblock {\em Int. J. Comput. Vis.}, vol. 24, no. 1, pp. 57--78, 1997.

\bibitem{sethian1999fast}
J.~A. Sethian,
\newblock ``Fast marching methods,''
\newblock {\em SIAM Review}, vol. 41, no. 2, pp. 199--235, 1999.

\bibitem{mirebeau2014anisotropic}
J.-M. Mirebeau,
\newblock ``Anisotropic fast-marching on cartesian grids using lattice basis
  reduction,''
\newblock {\em SIAM J. Numer. Anal.}, vol. 52, no. 4, pp. 1573--1599, 2014.

\bibitem{mirebeau2014efficient}
J.-M. Mirebeau,
\newblock ``{Efficient fast marching with Finsler metrics},''
\newblock {\em Numer. Math.}, vol. 126, no. 3, pp. 515--557, 2014.

\bibitem{benmansour2009fast}
F.~Benmansour and L.~D. Cohen,
\newblock ``{Fast object segmentation by growing minimal paths from a single
  point on 2D or 3D images},''
\newblock {\em J. Math. Imag. Vis.}, vol. 33, no. 2, pp. 209--221, 2009.

\bibitem{kaul2012detecting}
V.~Kaul, A.~Yezzi, and Y.~Tsai,
\newblock ``Detecting curves with unknown endpoints and arbitrary topology
  using minimal paths,''
\newblock {\em IEEE Trans. Pattern Anal. Mach. Intell.}, vol. 34, no. 10, pp.
  1952--1965, 2012.

\bibitem{li20093d}
H.~Li, A.~Yezzi, and L.~D. Cohen,
\newblock ``{3D multi-branch tubular surface and centerline extraction with 4D
  iterative key points},''
\newblock in {\em Proc. MICCAI}, 2009, pp. 1042--1050.

\bibitem{chen2016vesselkeypoints}
D.~Chen, J.-M. Mirebeau, and L.~D. Cohen,
\newblock ``Vessel tree extraction using radius-lifted keypoints searching
  scheme and anisotropic fast marching method,''
\newblock {\em J. Algorithm Comput. Technol.}, vol. 10, no. 4, pp. 224--234,
  2016.

\bibitem{rouchdy2013geodesic}
Y.~Rouchdy and L.~D. Cohen,
\newblock ``{Geodesic voting for the automatic extraction of tree structures.
  Methods and applications},''
\newblock {\em Comput. Vis. Image Understand.}, vol. 117, no. 10, pp.
  1453--1467, 2013.

\bibitem{mille2009deformable}
J.~Mille and L.~D. Cohen,
\newblock ``{Deformable tree models for 2D and 3D branching structures
  extraction},''
\newblock in {\em Proc. CVPRW}, 2009, pp. 149--156.

\bibitem{chen2016curve}
Y.~Chen et~al.,
\newblock ``Curve-like structure extraction using minimal path propagation with
  backtracking,''
\newblock {\em IEEE Trans. Image Process.}, vol. 25, no. 2, pp. 988--1003,
  2016.

\bibitem{cohen2001grouping}
L.~D. Cohen and T.~Deschamps,
\newblock ``Grouping connected components using minimal path techniques.
  application to reconstruction of vessels in 2d and 3d images,''
\newblock in {\em Proc. CVPR}, 2001, vol.~2.

\bibitem{deschamps2001fast}
T.~Deschamps and L.~D. Cohen,
\newblock ``{Fast extraction of minimal paths in 3D images and applications to
  virtual endoscopy},''
\newblock {\em Med. Image Anal.}, vol. 5, no. 4, pp. 281--299, 2001.

\bibitem{li2007vessels}
H.~Li and A.~Yezzi,
\newblock ``{Vessels as 4-D curves: Global minimal 4-D paths to extract 3-D
  tubular surfaces and centerlines},''
\newblock {\em IEEE Trans. Med. Imag.}, vol. 26, no. 9, pp. 1213--1223, 2007.

\bibitem{benmansour2011tubular}
F.~Benmansour and L.~D. Cohen,
\newblock ``Tubular structure segmentation based on minimal path method and
  anisotropic enhancement,''
\newblock {\em Int. J. Comput. Vis.}, vol. 92, no. 2, pp. 192--210, 2011.

\bibitem{pechaud2009extraction}
M.~P{\'e}chaud, R.~Keriven, and G.~Peyr{\'e},
\newblock ``Extraction of tubular structures over an orientation domain,''
\newblock in {\em Proc. CVPR}, 2009, pp. 336--342.

\bibitem{chen2017global}
D.~Chen, J.-M. Mirebeau, and L.~D. Cohen,
\newblock ``{Global minimum for a Finsler elastica minimal path approach},''
\newblock {\em Int. J. Comput. Vis.}, vol. 122, no. 3, pp. 458--483, 2017.

\bibitem{bekkers2015pde}
E.~J. Bekkers, R.~Duits, A.~Mashtakov, and G.~R. Sanguinetti,
\newblock ``{A PDE approach to data-driven sub-Riemannian geodesics in
  SE(2)},''
\newblock {\em SIAM J. Imag. Sci.}, vol. 8, no. 4, pp. 2740--2770, 2015.

\bibitem{mashtakov2017tracking}
A.~Mashtakov et~al.,
\newblock ``{Tracking of lines in spherical images via sub-Riemannian geodesics
  in SO(3)},''
\newblock {\em J. Math. Imag. Vis.}, vol. 58, no. 2, pp. 239--264, 2017.

\bibitem{duits2018optimal}
R.~Duits et~al.,
\newblock ``{Optimal paths for variants of the 2D and 3D Reeds--Shepp car with
  applications in image analysis},''
\newblock {\em J. Math. Imag. Vis.}, vol. 60, no. 6, pp. 816--848, 2018.

\bibitem{mirebeau2018fast}
J.-M. Mirebeau,
\newblock ``Fast-marching methods for curvature penalized shortest paths,''
\newblock {\em J. Math. Imag. Vis.}, vol. 60, no. 6, pp. 784--815, 2018.

\bibitem{liao2018progressive}
W.~Liao et~al.,
\newblock ``{Progressive minimal path method for segmentation of 2D and 3D line
  structures},''
\newblock {\em IEEE Trans. Pattern Anal. Mach. Intell.}, vol. 40, no. 3, pp.
  696--709, 2018.

\bibitem{chen2015interactive}
D.~Chen and L.~D. Cohen,
\newblock ``Interactive retinal vessel centreline extraction and boundary
  delineation using anisotropic fast marching and intensities consistency,''
\newblock in {\em Proc. EMBC}, 2015, pp. 4347--4350.

\bibitem{geusebroek2003fast}
J.-M. Geusebroek, A.~W. Smeulders, and J.~Van De~Weijer,
\newblock ``Fast anisotropic gauss filtering,''
\newblock {\em IEEE Trans. Image Process.}, vol. 12, no. 8, pp. 938--943, 2003.

\bibitem{franceschiello2018neuro}
B.~Franceschiello, A.~Sarti, and G.~Citti,
\newblock ``{A Neuromathematical model for geometrical optical illusions},''
\newblock {\em J. Math. Imag. Vis.}, vol. 60, no. 1, pp. 94--108, 2018.

\bibitem{tsitsiklis1995efficient}
J.~N. Tsitsiklis,
\newblock ``Efficient algorithms for globally optimal trajectories,''
\newblock {\em IEEE Trans. Automat. Contr.}, vol. 40, no. 9, pp. 1528--1538,
  1995.

\bibitem{chen2015piecewise}
D.~Chen and L.~D. Cohen,
\newblock ``Piecewise geodesics for vessel centerline extraction and boundary
  delineation with application to retina segmentation,''
\newblock in {\em Proc. SSVM}, 2015, pp. 270--281.

\bibitem{wu2007leaf}
S.~G. Wu et~al.,
\newblock ``A leaf recognition algorithm for plant classification using
  probabilistic neural network,''
\newblock in {\em Proc. IEEE Int. Symp. Signal Proc. Inf. Tech.}, 2007, pp.
  11--16.

\bibitem{staal2004ridge}
J.~Staal et~al.,
\newblock ``Ridge-based vessel segmentation in color images of the retina,''
\newblock {\em IEEE Trans. Med. Imag.}, vol. 23, no. 4, pp. 501--509, 2004.

\bibitem{hu2013automated}
Q.~Hu, M.~D. Abr{\`a}moff, and M.~K. Garvin,
\newblock ``Automated separation of binary overlapping trees in low-contrast
  color retinal images,''
\newblock in {\em Proc. MICCAI}, 2013, pp. 436--443.

\end{thebibliography}

\begin{IEEEbiographynophoto}{Da Chen} received his Ph.D degree in applied mathematics  from CEREMADE, University Paris Dauphine, PSL Research University,  Paris, France, in 2016, supervised by Professor Laurent D. Cohen. Currently he is a postdoc researcher  at CEREMADE, University Paris Dauphine, PSL Research University, and also at Centre Hospitalier National d’Ophtalmologie des Quinze-Vingts, Paris, France. His research interests include variational methods, partial differential equations, and geometry methods as well as their applications to medical image analysis, such as image segmentation, tubular structure extraction and medical image registration. 
\end{IEEEbiographynophoto}

\begin{IEEEbiographynophoto}{Jiong Zhang} received the master’s degree in computer science from the Northwest A\&F University, Yangling, China, and the Ph.D. degree from the Eindhoven University of Technology, The Netherlands. He then joined as a Post-Doctoral Researcher with the Medical Image Analysis Group, Eindhoven University of Technology, The Netherlands. Currently, he is working as a Postdoctoral researcher in Laboratory of Neuro Imaging (LONI), Keck School of Medicine of University of Southern California. His research interests include ophthalmologic image analysis, medical image analysis, computer-aided diagnosis, and machine learning. 
\end{IEEEbiographynophoto}

\begin{IEEEbiographynophoto}{Laurent D. Cohen} was student at the Ecole Normale Superieure, rue d'Ulm in Paris, France, from 1981 to 1985. He received the Master's and Ph.D. degrees in applied mathematics from University of Paris 6, France, in 1983 and 1986, respectively. He got the Habilitation diriger des Recherches from University Paris $9$ Dauphine in 1995. From 1985 to 1987, he was member at the computer graphics and image processing group at Schlumberger Palo Alto Research, Palo Alto, California, and Schlumberger Montrouge Research, Montrouge, France, and remained consultant with Schlumberger afterward. He began working with INRIA, France, in 1988, mainly with the medical image understanding group EPIDAURE. He obtained in 1990 a position of Research Scholar (Charge then Directeur de Recherche 1st class) with the French National Center for Scientific Research (CNRS) in the Applied Mathematics and Image Processing group at CEREMADE, Universite Paris Dauphine, Paris, France. His research interests and teaching at university are applications of partial differential equations and variational methods to image processing and computer vision, such as deformable models, minimal paths, geodesic curves, surface reconstruction, image segmentation, registration and restoration. He is currently or has been editorial member of the Journal of Mathematical Imaging and Vision, Medical Image Analysis and Machine Vision and Applications. He was also member of the program committee for about 50 international conferences. He has authored about 260 publications in international Journals and conferences or book chapters, and has 6 patents. In 2002, he got the CS 2002 prize for Signal and Image Processing. In 2006, he got the Taylor \& Francis Prize: ``2006 prize for Outstanding innovation in computer methods in biomechanics and biomedical engineering." He was 2009 laureate of Grand Prix EADS de l’Academie des Sciences in France. He was promoted as IEEE Fellow 2010 for contributions to computer vision technology for medical imaging.
\end{IEEEbiographynophoto}

\end{document}